\documentclass[10pt,journal,compsoc,manuscript]{IEEEtran}

\ifCLASSOPTIONcompsoc
  \usepackage[nocompress]{cite}
\else
  \usepackage{cite}
\fi

\usepackage{cite}
\usepackage{amsmath,amssymb,amsfonts}
\usepackage{algorithm}
\usepackage[noend]{algpseudocode}
\usepackage{wrapfig, blindtext}
\usepackage[export]{adjustbox}
\usepackage{graphicx}
\usepackage{textcomp}
\usepackage{xcolor}

\usepackage[utf8]{inputenc} 
\usepackage[T1]{fontenc}    
\usepackage{hyperref}       
\usepackage{url}            
\usepackage{booktabs}       
\usepackage{amsfonts}       
\usepackage{nicefrac}       
\usepackage{microtype}      
\usepackage{xcolor}         

\usepackage{diagbox}
\usepackage{mathtools}
\usepackage{amsmath}
\usepackage{graphicx}
\usepackage{pifont}
\usepackage{multirow}
\usepackage{multicol}
\usepackage{threeparttable}
\usepackage{comment}
\usepackage{subfigure}
\usepackage{wrapfig, blindtext}
\usepackage[export]{adjustbox}

\usepackage{contour}
\usepackage{ulem}

\contourlength{0.8pt}

\newcommand{\ul}[1]{%
  \uline{\phantom{#1}}%
  \llap{\contour{white}{#1}}%
}

\begin{document}
%
\title{Handling Inter-class and Intra-class Imbalance in Class-imbalanced Learning}
%
%
%
%

\author{
  Zhining~Liu,
  Pengfei~Wei,
  Zhepei~Wei,
  Boyang~Yu,
  Yuan~Tian,
  Jing~Jiang,
  Wei~Cao,
  Jiang~Bian,
  Yi~Chang
\IEEEcompsocitemizethanks{
\IEEEcompsocthanksitem Z. Liu is with University of Illinois Urbana-Champaign, USA. 
Email: liu326@illinois.edu.
P. Wei is with ByteDance, Singapore.
Email: pengfei.wei@bytedance.com.
Z. Wei is with University of Virginia, USA. Email: tqf5qb@virginia.edu.
B. Yu is with JD.com, China. Email: yuboyang9@jd.com.
Y. Tian and Y. Chang are with Jilin University, China.
Email: \{yuantian, yichang\}@jlu.edu.cn.
J. Jiang is with University of Technology Sydney, Australia.
Email: jing.jiang@uts.edu.au.
W. Cao and J. Bian are with Microsoft Research.
Email: \{weicao,jiang.bian\}@microsoft.com.
\IEEEcompsocthanksitem Yi Chang is the corresponding author.
This work was mainly done while Z. Liu, Z. Wei and B. Yu were studying at Jilin University.
}
}

%
%

\markboth{Journal of \LaTeX\ Class Files,~Vol.~14, No.~8, August~2015}%
{Shell \MakeLowercase{\textit{et al.}}: Bare Demo of IEEEtran.cls for Computer Society Journals}
%


\newcommand{\method}{{\sc DuBE} }
\newcommand{\codeurl}{https://github.com/AnonAuthorAI/duplebalance}
\newcommand{\coderef}{\href{\codeurl}{\underline{Github}}}
\newcommand{\docurl}{https://duplebalance.readthedocs.io}
\newcommand{\docref}{\href{\docurl}{\underline{ReadtheDocs}}}
\newcommand{\yes}{\ding{51}}
\newcommand{\no}{\ding{55}}
\newcommand{\yesno}{\textcolor{black}{\ding{51}}{\textcolor{black}{\kern-0.65em\ding{55}}}}

\newcommand{\argmax}{\operatornamewithlimits{argmax}}
\newcommand{\argmin}{\operatornamewithlimits{argmin}}


\IEEEtitleabstractindextext{%
\begin{abstract}
Class-imbalance is a common problem in machine learning practice. 
Typical Imbalanced Learning (IL) methods balance the data via intuitive class-wise resampling or reweighting. 
However, previous studies suggest that beyond class-imbalance, intrinsic data difficulty factors like overlapping, noise, and small disjuncts also play critical roles. 
To handle them, many solutions have been proposed (e.g., noise removal, borderline sampling, hard example mining) but are still confined to a specific factor and cannot generalize to broader scenarios, which raises an interesting question: how to handle both the class-agnostic difficulties and the class-imbalance in a unified way? 
To answer this, we consider both class-imbalance and its orthogonal: intra-class imbalance, i.e., the imbalanced distribution over easy and hard samples. 
Such distribution naturally reflects the complex influence of class-agnostic intrinsic data difficulties thus providing a new unified view for identifying and handling these factors during learning. 
From this perspective, we discuss the pros and cons of existing IL solutions and further propose new balancing techniques for more robust and efficient IL. 
Finally, we wrap up all solutions into a generic ensemble IL framework, namely DuBE (Duple-Balanced Ensemble). 
It features explicit and efficient inter-\&intra-class balancing as well as easy extension with standardized APIs. Extensive experiments validate the effectiveness of DuBE. 
Code, examples, and documentation are available at {\color{blue}\coderef}\footnote{\href{\codeurl}{\underline{\tt \codeurl}}} and {\color{blue}\docref}\footnote{\href{\docurl}{\underline{\tt \docurl}}}.
\end{abstract}

}

\maketitle

\IEEEdisplaynontitleabstractindextext

%
\IEEEpeerreviewmaketitle

\IEEEraisesectionheading{\section{Introduction}\label{sec:introduction}}

\IEEEPARstart{M}{ost} of well-known machine learning algorithms work under the balanced sample assumption where the training samples are approximately evenly distributed over classes~\cite{he2008overview}.
However, this assumption does not always hold in practice.
Due to the naturally-skewed class distributions, class-imbalance has been widely observed in many real-world application domains including computer vision, fraud detection, medical diagnosis, etc~\cite{haixiang2017learning-from-imb-review,he2013overview,albert02018experiment}.
Facing the class-imbalance, canonical learning models usually suffer from the "majority bias", i.e., performing well in majority classes but poorly in minority classes.
Such a result is suboptimal as the minority classes are typically of primary interest~\cite{he2013overview}.

\newcommand{\introexamplewidth}{0.45\linewidth}

\begin{figure}[t]
  \centering
  \subfigure[Class distribution]{
    \includegraphics[width=\introexamplewidth]{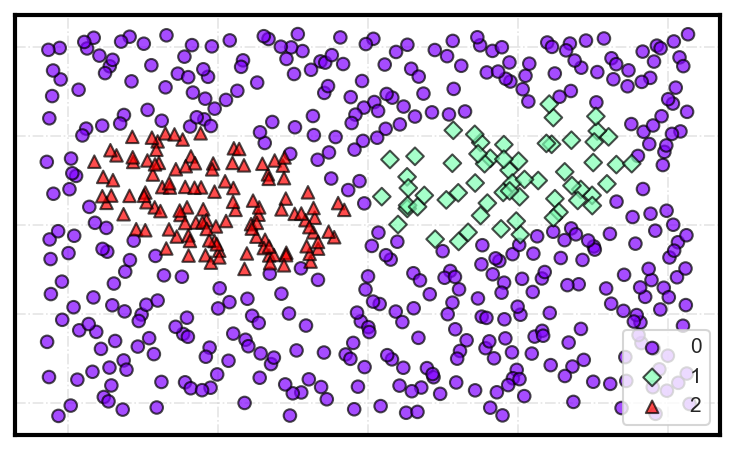}
    \label{fig:example-0}
  }
  \subfigure[Inter-class imbalance]{
    \label{fig:example-1}
    \includegraphics[width=\introexamplewidth]{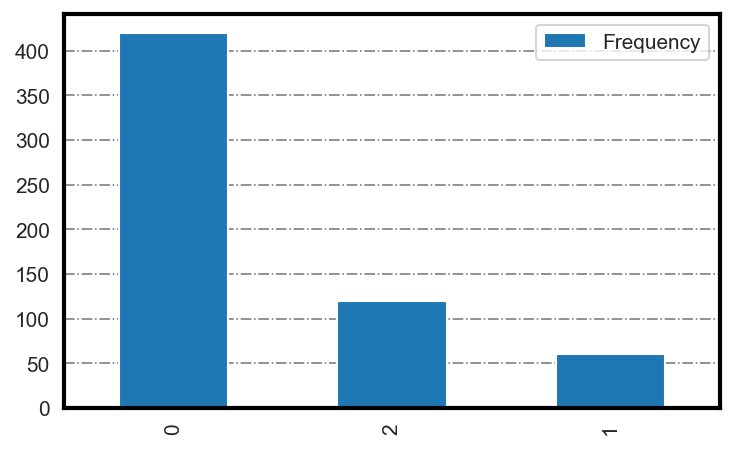}
  }
  \subfigure[Error distribution]{
    \label{fig:example-2}
    \includegraphics[width=\introexamplewidth]{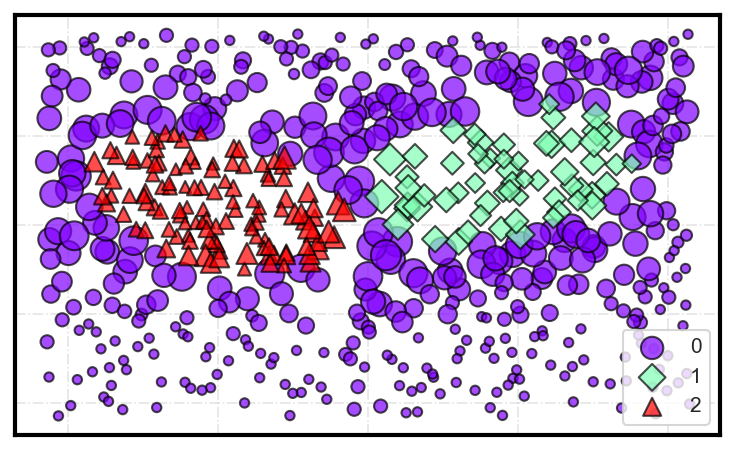}
  }
  \subfigure[Intra-class imbalance]{
    \label{fig:example-3}
    \includegraphics[width=\introexamplewidth]{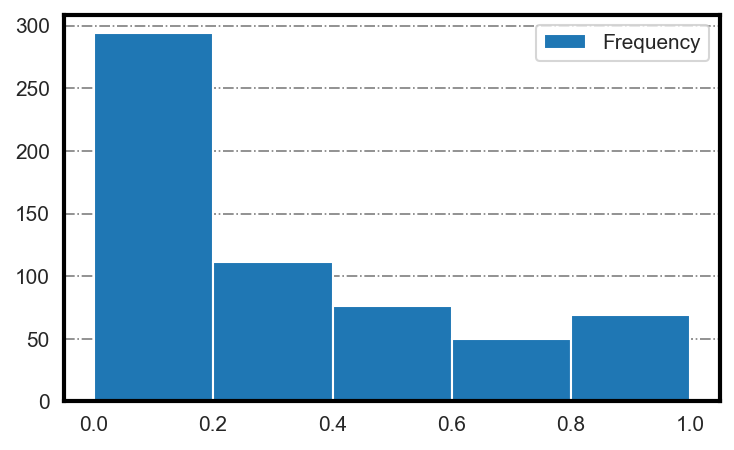}
  }
  \caption{
    An illustrative example of inter-class and intra-class imbalance, best viewed in color.
    Fig. \ref{sub@fig:example-0}: an imbalanced 3-class toy dataset.
    Fig. \ref{sub@fig:example-1}: the inter-class imbalanced distribution (of the number of samples).
    Fig. \ref{sub@fig:example-2}: the dataset resized by the prediction error of a classifier.
    Fig. \ref{sub@fig:example-3}: the intra-class imbalanced distribution (of the prediction error).
  }
  \label{fig:example}
\end{figure}

Imbalanced Learning (IL) aims to eliminate the "majority bias", it has long been an active area of machine learning and data mining research.
Most of the existing IL solutions are developed under an intuitively appealing strategy, which is to \textit{reduce class-imbalance by class-wise resampling or reweighting}.
This can be achieved by over-sampling minority classes (e.g.,~\cite{chawla2002smote,he2008adasyn,han2005borderline-smote}), under-sampling majority classes (e.g.,~\cite{mani2003nearmiss,tomek1976tomeklink,wilson1972enn,kubat1997oss}), or assigning different mis-classification costs to different classes (e.g.,~\cite{liu2006cost-sensitive-imbalance,cui2019cbloss,ling2004csdt,chai2004csnb}).
Nevertheless, some recent works arise under a new strategy without any class-level operation.
They \textit{implicitly facilitate class balance through instance-wise rectification} and have achieved notable success in certain areas such as vision-based classification and detection (e.g.,~\cite{shrivastava2016ohem,lin2017focalloss,liu2020mesa,liu2020self-paced-ensemble,dong2017rectification-hem,pang2019libra}).
The fact that class imbalance bias be alleviated without class-wise manipulation prompts us to think about some interesting questions:
\textit{\textbf{(i)}} Beyond the cardinality difference (imbalance) between classes, how does the difference between samples, even within the same class, affect learning?
\textit{\textbf{(ii)}} What difficulties or factors in the data cause such an intra-class difference?
\textit{\textbf{(iii)}} Finally, can we find a unified view to describe and handle the two orthogonal (inter-class \& intra-class) differences?



\begin{figure}[t]
  \centering
  \includegraphics[width=1\linewidth]{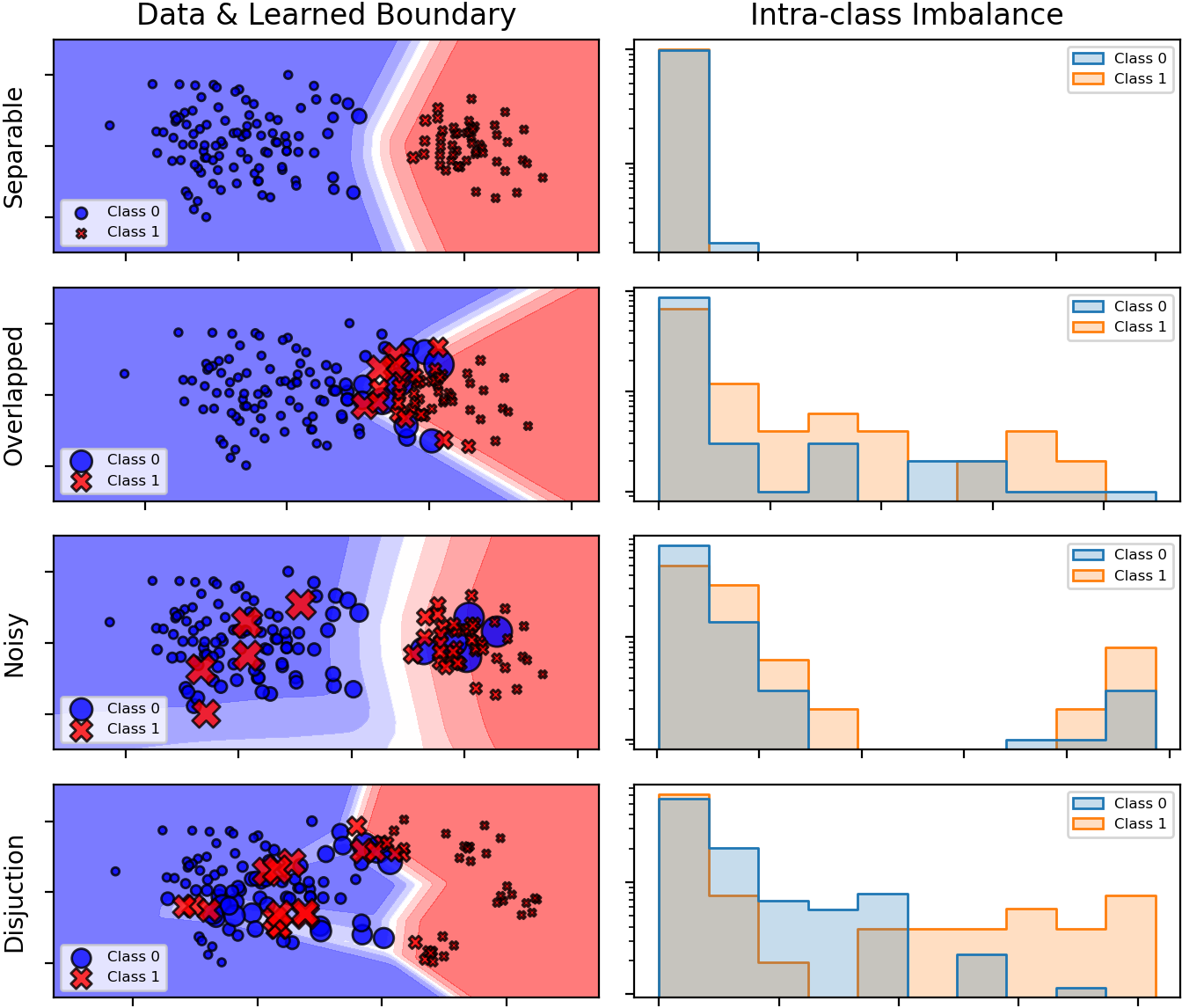}
  \centering
  \caption{
    Intra-class imbalance indicates the existence of data difficulty factors.
    In each row, the left side shows a class-imbalanced dataset with a specific factor and the decision boundary given by an MLPClassifier.
    We enlarged the data points with high prediction errors for better visualization.
    The right side shows the error distribution (intra-class imbalance) for each class.
    We observe that intra-class imbalance reflects the unique patterns of different difficulty factors:
    \textbf{1st row}: linearly separable data is easy to classify and has no intra-class imbalance;
    \textbf{2nd row}: overlapping makes instances located near the boundary hard to classify, causing long-tailed error distribution for both classes;
    \textbf{3rd row}: noise examples can barely be learned by the model, thus has a high error and result in "U"-shaped error distribution;
    \textbf{4th row}: minority disjunctions located in majority areas form a large tail in the minority class error distribution.
    }
  \label{fig:difffactor}
\end{figure}

In this paper, we answer the above questions by considering two kinds of imbalance presented in IL, i.e., \textbf{inter-class imbalance} and \textbf{intra-class imbalance}. 
\figurename~\ref{fig:example} gives illustrative examples of the two types of imbalance that co-exist in a dataset.
As can be readily observed, \figurename~\ref{fig:example-0} and \figurename~\ref{fig:example-1} show the inter-class imbalance, i.e., the uneven sample distribution between classes. 
Beyond this, \figurename~\ref{fig:example-2} and \figurename~\ref{fig:example-3} depict another imbalance that is orthogonal to the class-imbalance, namely intra-class imbalance, i.e., the uneven distribution of the quantity of easy and hard examples.
We note that despite its conceptual simplicity, \textit{intra-class imbalance can be an important indicator of data complexity, thus guiding the model to exploit supervision signals in a more efficient manner}.
For example, many previous works discussed the influence of intrinsic data difficulty factors (DDFs) in imbalanced learning, including class overlapping~\cite{napierala2016types-minority-samples,denil2010overlap,garcia2007overlap}, the presence of noise~\cite{garcia2015noise,koziarski2020multi-ccr}, and the decomposition of small disjuncts~\cite{prati2004overlap-small-disjuncts,jo2004small-disjuncts,garcia2015noise-small-disjuncts}.
Despite their heterogeneity, the intra-class imbalance can serve as an indicator of the existence of these DDFs, as shown in Fig.~\ref{fig:difffactor}.
We can see that intra-class imbalance not only reflects the existence of the DDFs, but also captures the unique patterns of different DDFs (please refer to the caption of Fig.~\ref{fig:difffactor} for more details), which makes it a powerful tool for describing the class-agnostic data complexities in class-imbalanced learning problems.

We note that there are related works that attempt to handle the instance-wise difference in IL from two different perspectives: 
\textit{\textbf{(i)}} some focus on handling specific data difficulty factor(s) by e.g., borderline instance generation~\cite{han2005borderline-smote,he2008adasyn}, noise removal~\cite{ramentol2012smotersb,napierala2010learn-from-noisy-borderline-data}, and clustering-based sampling~\cite{lin2017cus}.
However, it is hard to identify a specific DDF in complex real-world datasets, and multiple DDFs are highly likely to co-exist. 
The intuitive designs of these methods cannot generalize and well handle tasks that do not fit their assumptions on DDF.
Moreover, most of them rely heavily on neighborhood computing (e.g.,~\cite{han2005borderline-smote,ramentol2012smotersb,batista2004smoteenn,he2008adasyn}), which requires a well-defined distance metric and up to $O(n^2)$ computational overhead. 
\textit{\textbf{(ii)}} Other works focus on mining hard examples, which they consider more informative and need to be emphasized during training~\cite{liu2009ee-bc,shrivastava2016ohem}.
However, this could make the model susceptible to noise and outliers as they will also have large errors and be wrongly emphasized in training (e.g., 3rd row of Fig.~\ref{fig:difffactor}).
Besides, recent works are mostly designed for training deep neural networks~\cite{shrivastava2016ohem,lin2017focalloss,dong2017rectification-hem} thus lacking the versatility to other learning models and tasks.
More importantly, no existing work explicitly considers the inter-class and intra-class imbalance and works on them.

In light of this, we propose DuBE (\underline{Du}ple-\underline{B}alanced \underline{E}nsemble) to handle the IL from the perspective of inter-class and intra-class imbalance.
As described above, the two concepts depict the two orthogonal imbalances that co-exist in imbalanced datasets, where the inter-class imbalance refers to the skewed class distribution, and the intra-class imbalance provides a unified view of the class-agnostic data difficulty factors and task complexity.
For inter-class balancing, we find that balanced under-sampling is an unbiased correction for skewed $P(Y)$ but is unstable due to information loss, whereas the opposite is true for over-sampling.
However, we further find that over-sampling can benefit more from proper data augmentation, thus hybrid-sampling and adaptive data augmentation mechanisms are introduced for more accurate and stable inter-class balancing.
For intra-class balancing, we discuss why naive hard example mining cannot handle noise/outliers and propose a simple and effective balancing strategy for fast and robust hard example mining.
Finally, we wrap all balancing strategies into DuBE and verify their effectiveness through extensive experiments and analysis.
We emphasize that DuBE is a versatile learning framework that performs explicit inter-\&intra-class balancing with modularized design, new balancing strategies can be easily integrated using its standard APIs.

\begin{figure*}[t]
  \centering
  \includegraphics[width=0.98\linewidth]{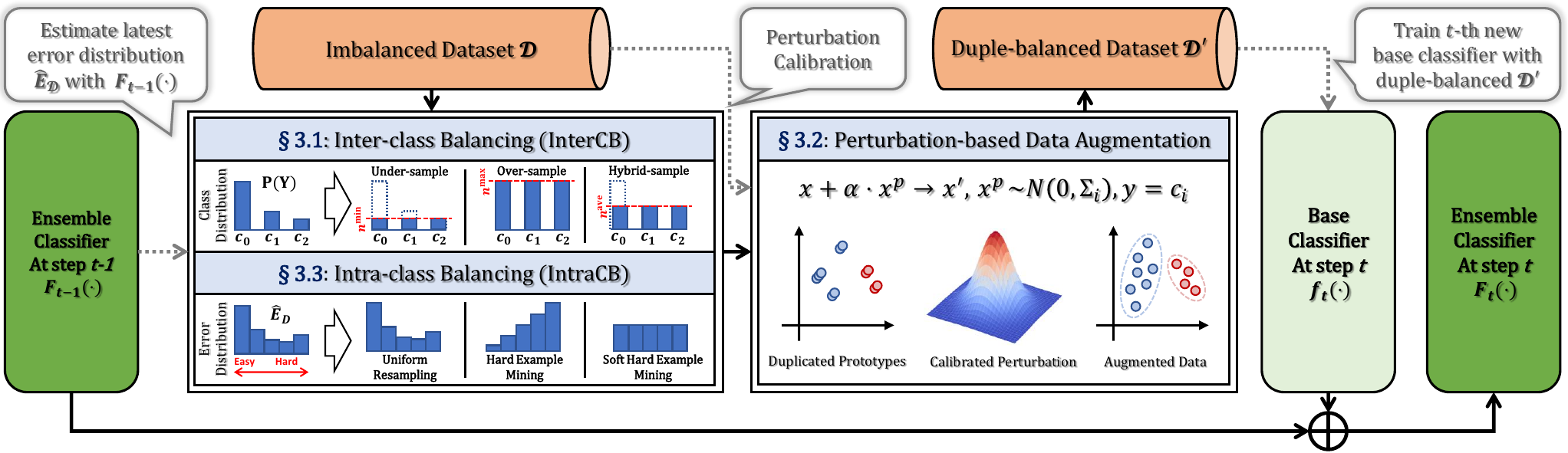}
  \centering
  \caption{Overview of the proposed \method Framework. Best viewed in color.}
  \label{fig:framework}
\end{figure*}

To sum up, this paper makes the following contributions:
\begin{itemize}
    \item To our best knowledge, we carry out the first explicit preliminary exploration of the inter-class and intra-class imbalance in IL problems. We discuss how the two types of imbalance are connected with intrinsic data difficulty factors as well as existing IL solutions.
    \item We present detailed discussions and analyses of the pros and cons of different IL solutions, based on this, we further proposed several techniques for fast and robust inter-class and intra-class balancing.
    \item We propose DuBE, a generic, efficient, and modularized ensemble imbalanced learning framework. It features explicit inter-\&intra-class balancing as well as easy integration, combination, and extension of balancing algorithms. Extensive experiments validate the effectiveness of DuBE.
\end{itemize}

The rest of this paper is organized as follows:
Section~\ref{sec:preliminaries} introduces the notations and the problem definition.
Section~\ref{sec:methodology} describes the \method framework, along with detailed analyses of the included inter/intra class balancing strategies.
Section~\ref{sec:experiments} presents experimental results and related discussions. 
Section~\ref{sec:related-works} reviews the related works.
And finally, Section~\ref{sec:conclusion} concludes the paper.
\section{Preliminaries}
\label{sec:preliminaries}


Before getting into the \method framework, we first briefly introduce the notations used in this paper and formally define the class-imbalanced learning problem in this section.

\begin{table}[h]
\caption{Definitions of basic notations}
\label{tab:notation}
\centering
\begin{tabular}{c|c}
\toprule
\textbf{Notation}                                & \textbf{Definition}                            \\ \midrule
$d$                                              & Input dimensionality.                          \\ \hline
$m$                                              & Number of classes.                             \\ \hline
$N$                                              & Number of data instances.                      \\ \hline
$\mathcal{X}:\mathbb{R}^d$                       & Input feature space.                           \\ \hline
$\mathcal{Y}:\{c_1, c_2, ..., c_m\}$             & Output label space.                            \\ \hline
$s:(x, y)$                                       & Data instance, where $x \in \mathcal{X}, y \in \mathcal{Y}$. \\ \hline
$D:\{s_i\}_{i=1, 2, ..., N}$                     & Dataset with $N$ examples.                     \\ \hline
$D_c:\{(x, y)|y=c\}$                             & Subset of data examples from class $c$.        \\ \hline
$F: \mathcal{X} \to \mathcal{Y}$                 & A learning-based classifier.                   \\ \hline
$f: \mathcal{X} \to \mathcal{Y}$                 & Base learner of an ensemble classifier.        \\ \hline
$F_k: \mathcal{X} \to \mathcal{Y}$               & An ensemble classifier with $k$ base learners. \\ \bottomrule
\end{tabular}
\end{table}

\textbf{Notations.}
Formally, let $N$ denote the number of instances in the dataset and $d$ be the input dimensionality, i.e., the number of input features.
We can define the feature space $\mathcal{X}:\mathbb{R}^d$ and label space $\mathcal{Y}:\{c_1, c_2, ..., c_m\}$, where $m$ is the number of classes.
Then a data sample is $s:(x, y)$ with $x \in \mathcal{X}, y \in \mathcal{Y}$.
Likewise, a dataset $D$ can be represented as $D:\{s_i\}_{i=1, 2, ..., N}$.
A classifier is represented by $F(\cdot)$, which is a projection from the feature space to the label space, i.e., $F: \mathcal{X} \to \mathcal{Y}$.
Moreover, we use $f(\cdot)$ to denote a base learner of an ensemble classifier, and $F_k$ as an ensemble classifier with $k$ base learners.
Table~\ref{tab:notation} summarizes the main notation definitions in this paper.

\textbf{Problem definition.}
Note that the most of the existing works are conducted based on the binary case of imbalanced learning, i.e., $m=2$~\cite{haixiang2017learning-from-imb-review,he2013overview,he2008overview}. 
Without loss of generality, we describe the binary IL problem in this section, and extension to the multi-class scenario is straightforward.
In this case, we consider the majority class $c_\text{maj}$ and the minority class $c_\text{min}$.
Let's denote the set of data examples from class $c$ as $D_c:\{(x, y)|y=c\}$.
Then the class-imbalance refers to the fact that $|D_{c_\text{maj}}| \gg |D_{c_\text{min}}|$.
That is, the examples are not evenly distributed over different classes, i.e., the underlying class (marginal) distribution \textbf{P}(Y) is skewed.
Under such class-imbalance, the learning process of canonical machine learning methods are likely to be dominant by the majority class examples due to the accuracy-oriented learning objectives~\cite{he2008overview}. 
This usually leads to poor prediction performance for minority classes.
Therefore, \textit{the goal of imbalanced learning (IL) is to learn an unbiased classifier $F: \mathcal{X} \to \mathcal{Y}$ from a skewed dataset $D$}.
With the above notations and definitions, we now present our \method framework and the associated analysis.



\section{Analysis \& Methodology}
\label{sec:methodology}





In this section, we present a detailed discussion on the inter-class and intra-class balancing strategies considered in this paper. 
We demonstrate their pros and cons in detail, and introduce several new techniques for more accurate and robust balanced learning.
Specifically, we answer following {\bf Research Questions (RQs)} in each subsection:
\begin{itemize}
    \item \textbf{\textit{RQ1:}} 
    What are the advantages and disadvantages of different inter-class rebalancing strategies (e.g., under/over-sampling)? 
    ($\triangleright$ Section~\ref{sec:intercb})
    \item \textbf{\textit{RQ2:}} 
    How can data augmentation help reduce decision bias in imbalanced learning?
    ($\triangleright$ Section~\ref{sec:pbda})
    \item \textbf{\textit{RQ3:}} 
    Why is hard example mining susceptible to noise and outliers, and how can more robust mining rules be designed based on the concept of intra-class balancing?
    ($\triangleright$ Section~\ref{sec:intracb})
\end{itemize}

Note that all techniques discussed in this section are also components of the proposed DuBE (\underline{Du}ple-\underline{B}alanced \underline{E}nsemble) framework.
As previously described, DuBE executes explicit inter-class and intra-class balancing simultaneously.
This is achieved by performing duple-balanced resampling within iterative ensemble training.
Specifically, in $t$-th iteration, an inter-class balancing strategy determines the target distribution of class-wise resampling, and an intra-class balancing strategy assigns instance-wise sampling probabilities w.r.t. the current learning state (fitted ensemble model) $F_{t-1}(\cdot)$.
A new base classifier $f_t(\cdot)$ is then trained using the duple-balanced resampled dataset $D'$ and added to the ensemble to form $F_t(\cdot)$.
\figurename~\ref{fig:framework} shows an overview of DuBE.
We now cover the analysis and technical details.

\subsection{Inter-class Balancing (InterCB).}
\label{sec:intercb}

We first introduce the three InterCB strategies considered in {\sc DuBE}, i.e., under-sampling, over-sampling, and hybrid-sampling, and then discuss why they are effective for InterCB.

\textbf{(1)}
\ul{\it Under-sampling}: The class containing the fewest samples is considered the minority class $c_\text{min}$, and all others are treated as majority ones. 
The majority classes are then under-sampled until they are of the same size as $c_\text{min}$, i.e., $\forall c \in \mathcal{Y}, |D'_c| = |D_{c_\text{min}}|$, where $D'$ represents the resampled dataset.
\textbf{(2)}
\ul{\it Over-sampling}: Conversely, the class containing the largest number of samples is considered as the majority class $c_\text{maj}$. 
Other classes are over-sampled via instance duplication until they are of the same size as $c_\text{maj}$, i.e., $\forall c \in \mathcal{Y}, |D'_c| = |D_{c_\text{maj}}|$.
\textbf{(3)}
\ul{\it Hybrid-sampling}: We further consider a combination of the aforementioned strategies, namely hybrid-sampling.
Majority and minority classes are distinguished by whether they contain more than average number of samples.
Then, we under/over-sample the majority/minority classes to bring them to the same size, i.e., $\forall c \in \mathcal{Y}, |D'_c| = \Sigma^{m}_{i=1}|D_{c_i}|/m = \frac{|D|}{m}$.

\begin{figure}[h]
  \centering
  \includegraphics[width=1\linewidth]{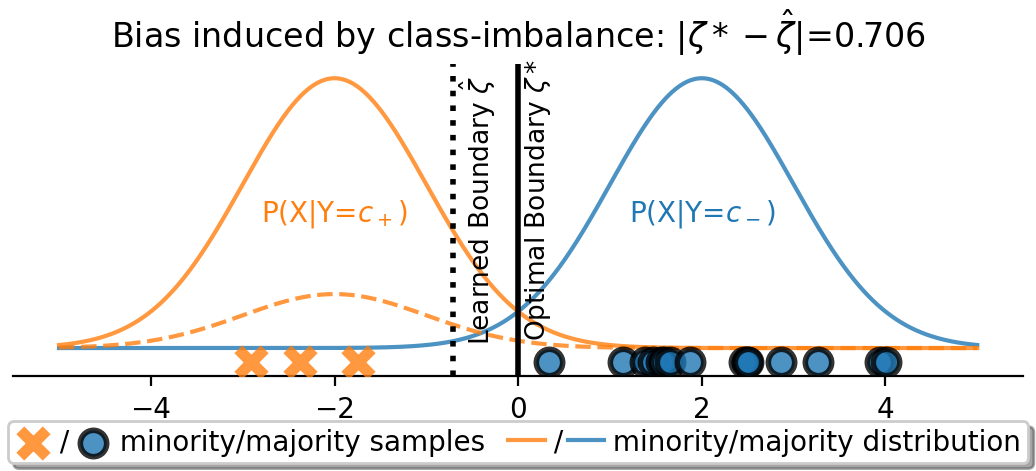}
  \caption{
  A toy imbalanced classification problem, best viewed in color.
  It contains 3 minority samples (the "$\times$"s) and 15 majority samples (the "$\circ$"s), i.e., imbalance ratio $|D_{c_\text{max}}|$/$|D_{c_\text{min}}|$ = 5.
  The underlying distributions of both classes are also included in the figure (colored solid lines), which are one-dimensional Gaussian distributions $\mathcal{N}(\mu, \sigma^2)$ with different $\mu$.
  The solid black line indicates the optimal decision boundary $\zeta^*$ with respect to the underlying distribution; and the dotted line represents the max-margin boundary $\hat{\zeta}$ on the sampled dataset.
  The bias induced by class-imbalance is defined as $|\zeta^*-\hat{\zeta}|$.
  }
  \label{fig:toy-data}
\end{figure}

\newcommand{\quadwidth}{0.23\linewidth}

\begin{figure*}[t]
  \centering
  \subfigure[Random Under-sampling.]{
    \includegraphics[width=\quadwidth]{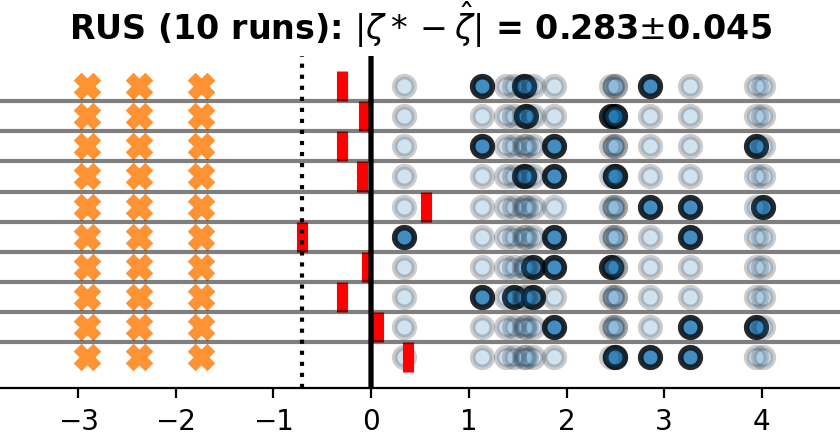}
    \label{fig:intercb-rus}
  }
  \subfigure[Random Over-sampling.]{
    \includegraphics[width=\quadwidth]{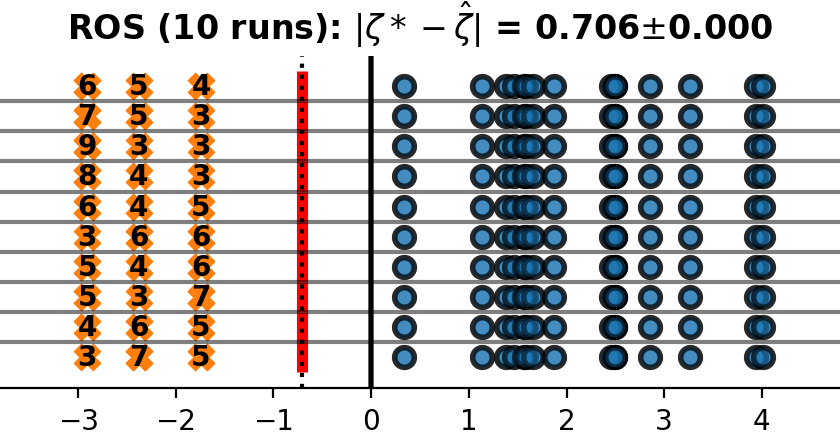}
    \label{fig:intercb-ros}
  }
  \subfigure[SMOTE Over-sampling~\cite{chawla2002smote}.]{
    \includegraphics[width=\quadwidth]{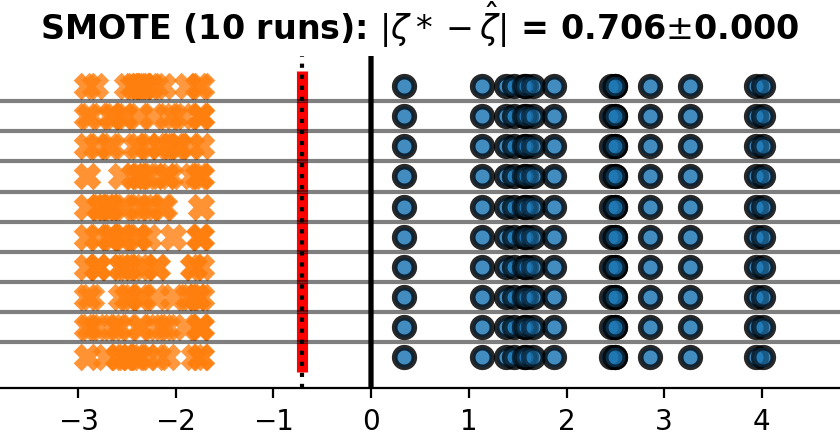}
    \label{fig:intercb-smote}
  }
  \subfigure[Random Hybrid-sampling.]{
    \includegraphics[width=\quadwidth]{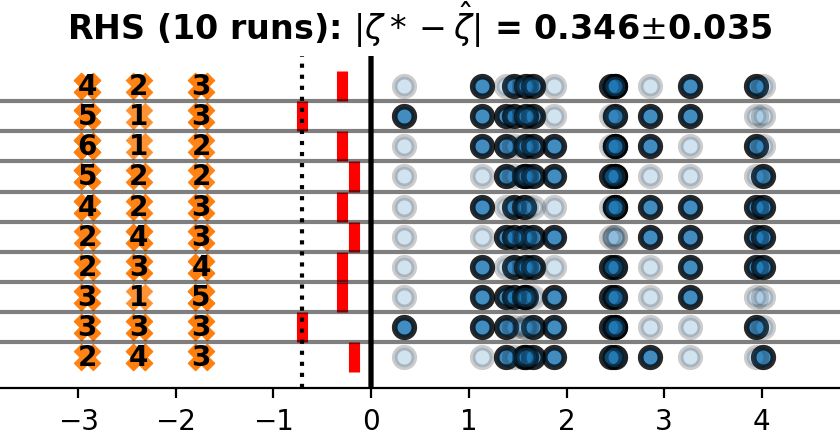}
    \label{fig:intercb-rhs}
  }
  \caption{
  A comparison of different inter-class balancing strategies. 
  Each subplot shows the resampling results of the corresponding method in 10 independent runs.
  The optimal boundary $\zeta^{*}$ (solid line), original learned $\hat{\zeta}_D$ (dotted line), and new boundaries $\hat{\zeta}_{D'}$ (red lines) learned from resampled $D'$ are also included.
  }
  \label{fig:intercb}
\end{figure*}

\newcommand{\cmin}{{c_+}}
\newcommand{\cmaj}{{c_-}}
\newcommand{\cmindist}{{\textit{P(X|Y=}\cmin\textit{)}}}
\newcommand{\cmajdist}{{\textit{P(X|Y=}\cmaj\textit{)}}}

\textbf{An example problem.}
In this section, we use an example to illustrate the effectiveness of the different InterCB strategies described above.
Please note that \textit{the following discussion is intended to provide an intuition about how different approaches can help IL, rather than a analysis of their theoretical properties.}
With this premise in mind, we consider a simple imbalanced classification setting, as shown in \figurename~\ref{fig:toy-data}.
It is a one-dimensional two-class imbalanced dataset, with 3 samples from the minority class and 15 samples from the majority class.
Without loss of generality, we assume the minority class to be positive, and denote the minority/majority class as $\cmin$/$\cmaj$. i.e., $c_\text{min}:=\cmin=1$ and $c_\text{maj}:=\cmaj=-1$.
The imbalance ratio (IR) is then $|D_\cmaj|/|D_\cmin| = 15/3 = 5$.
Each class is sampled from a normal (Gaussian) distribution $\mathcal{N}(\mu, \sigma^2)$, with the same $\sigma$=1 but different $\mu$, where $\mu_--\mu_+=4$.
It is obvious that the optimal decision boundary $\zeta^*$ with respect to the underlying distrbution $P(X|Y)$ should be $(\mu_++\mu_-)/2$ (the solid black line in \figurename~\ref{fig:toy-data}). 
That is, the optimal classifier predicts a sample to be positive if $x<(\mu_++\mu_-)/2$ and negative if otherwise.
However, the $\zeta^*$ cannot be learned from the sampled dataset due to the imbalanced marginal distribution $P(Y)$.
Let's consider a simple hard max-margin classifier $f(x):\text{sign}(x-b)$ that maximises the margin between the nearest points of opposite classes, which can be derived by solving:
\begin{equation}
\begin{aligned}
    \text{arg}&\text{max}_{\zeta} & & \gamma & \\
    & \text{s.t.}             & |x_i&-\zeta| \geq \gamma, &\ \forall (x_i, y_i) &\in D.\\
    &                         & -y_i(x_i&-\zeta) \geq 0,  &\ \forall (x_i, y_i) &\in D.\\
\end{aligned}
\end{equation}

The decision boundary $\hat{\zeta}$ of the max-margin classifier learned on the toy imbalanced dataset is represented by the dotted black line in \figurename~\ref{fig:toy-data}.
It can be seen that the learned $\hat{\zeta}$ is skewed towards the minority class compared with the optimal boundary $\zeta^*$.
This is due to the lack of support vector samples that are closer to the decision boundary in the minority class.
We can see that samples close to the decision boundary are rare patterns in the underlying distribution $P(X|Y)$ for both classes.
However, since the class distribution $P(Y)$ is skewed, the majority class is more likely to be well represented by the data and contains rare samples, which serve as stronger support vectors in the learning process (e.g., the leftmost majority sample in \figurename~\ref{fig:toy-data}).

Formally, for class $c$, let's use $d^{max}_{c}$ to denote the distance between the class distribution center $\mu_c$ and the support vector (the point farthest from $\mu_c$ in the sampled dataset), i.e., 
\begin{equation}
    d^{max}_c=\text{max}(|x-\mu_c|), 
    \forall (x, y) \in D_c, x\sim\mathcal{N}(\mu_c, 1).
    \label{eq2}
\end{equation}
Then the support vector of minority class can be written as $s_+^\text{sup}: (x_+^\text{sup}, \cmin)$ (e.g., the rightmost $\cmin$ instance in \figurename~\ref{fig:toy-data}), where $x_+^\text{sup} = \mu_++d^{max}_\cmin$.
Similarly, we have the majority support vector $s_-^\text{sup}: (x_-^\text{sup}, \cmaj), x_-^\text{sup} = \mu_--d^{max}_\cmaj$.
Then the max-margin separator $\hat{\zeta}=(x_+^\text{sup}+x_-^\text{sup})/2$.
Therefore, the expectation of decision bias is:
\begin{equation}
\begin{aligned}
    \mathbb{E}_\textit{P(XY)}[|\zeta^*-\hat{\zeta}|] & = \mathbb{E}_\textit{P(XY)}|[(x_+^\text{sup}+x_-^\text{sup})/2-(\mu_++\mu_-)/2]| \\
    & = 
    |[\mathbb{E}_\textit{P(XY)}(x_+^\text{sup}-\mu_+)+\mathbb{E}_\textit{P(XY)}(x_-^\text{sup}-\mu_-)]/2| \\
    & = 
    \left|\frac{
        \mathbb{E}_\cmindist[d^{max}_\cmin]-\mathbb{E}_\cmajdist[d^{max}_\cmaj]
    }{2}\right|
    \label{eq3}
\end{aligned}
\end{equation}
As the class distribution $P(Y)$ is skewed, we can expect the expectation of $\mathbb{E}_D[d^{max}_\cmaj]$ is larger than that of $\mathbb{E}_D[d^{max}_\cmin]$, i.e.,
\begin{equation}
    \mathbb{E}_D[d^{max}_\cmaj] > \mathbb{E}_D[d^{max}_\cmin], 
    \text{ if } |D_\cmaj| > |D_\cmin|.
    \label{eq4}
\end{equation}
Note that $\mathbb{E}_D[d^{max}]$ has no closed-form expression, but its lower and upper bound are linearly related to $\sqrt{\text{log}n}$~\cite{kamath2015bounds}.
Putting Eq.~\eqref{eq3} and Eq.~\eqref{eq4} together, we have $|D_\cmaj| > |D_\cmin| \Rightarrow \mathbb{E}_D[d^{max}_\cmaj] > \mathbb{E}_D[d^{max}_\cmin] \Rightarrow \mathbb{E}_D[|\zeta^*-\hat{\zeta}|] > 0$, which shows how class-imbalance can induce bias into the classifier learning.

We then apply the above IntraCB strategies to this example dataset to intuitively demonstrate their effectiveness in mitigating the decision bias $|\zeta^*-\hat{\zeta}|$ induced by class-imbalance.
The results are shown in \figurename~\ref{fig:intercb}.
Note that to minimize the effect of randomness, for each strategy, we show the results of 10 independent runs, including the resampling results, the new decision boundaries $\hat{\zeta}$ learned on the resampled datasets (red lines), and the corresponding bias (means$\pm$variance).

\textbf{Under-sampling.}
As shown in \figurename~\ref{fig:intercb-rus}, random under-sampling (RUS) is very effective in terms of reducing the bias towards the minority class.
On average, it greatly reduces the decision bias from 0.706 to 0.283.
The reason behind is that: \textit{randomly discarding instances from the majority class $\cmaj$ until $|D_\cmaj|=|D_\cmin|$} is equivalent to \textit{sampling only $|D_\cmin|$ examples from the majority underlying distribution}.
Therefore, RUS can be seen as an unbiased correction to the skewed marginal distribution $P(Y)$.
By reducing the $|D_\cmaj|$ to $|D'_\cmaj|=|D_\cmin|$, RUS equalizes the $\mathbb{E}_{D'}[d^{max}_\cmaj]$ and $\mathbb{E}_{D'}[d^{max}_\cmin]$ and thus lowers the expected bias $\mathbb{E}_{D'}[|\zeta^*-\hat{\zeta}|]$ on the resampled dataset $D'$.

On the downside, however, we note that the decision bias varies considerably over multiple runs (with the highest variance 0.045 among all competitors), indicating that the performance of RUS is not stable.
This is caused by an obvious reason: randomly discarding most of the majority examples can introduce significant information loss~\cite{krawczyk2016learning,he2013overview}, especially when the dataset is highly imbalanced.
As can be seen in \figurename~\ref{fig:intercb-rus}, while RUS can reduce the bias towards $\cmin$, it is sometimes too aggressive and even brings opposite bias towards $\cmaj$.
Intuitively, RUS is also contradict to the common practice in data mining, which is to collect more data and exploit all available information.

\textbf{Over-sampling and SMOTE.}
\figurename~\ref{fig:intercb-ros} shows the results of random over-sampling (ROS).
We can observe that, despite ROS also equalizes the $|D'_\cmaj|$ and $|D'_\cmin|$ in the resampled dataset $D'$ like RUS, it does not help in terms of mitigating the decision bias, i.e., $|\zeta^*-\hat{\zeta}_D| = |\zeta^*-\hat{\zeta}_{D'}|$.
The key difference between ROS and RUS is the \textit{equivalence of "resampling" and "sampling from the original distribution" for the target class}.
As stated above, the majority class set after RUS is equivalent to sampling from the original distribution $P(X|Y=\cmaj)$, except that it contains less samples.
However, this is not the case for ROS.
ROS performs over-sampling by replicating existing minority class instances (indicated by the numbers on minority samples in \figurename~\ref{fig:intercb-ros}).
It means that the new synthetic instances are sampled from a uniform distribution over existing minority examples $D_\cmin$, rather than from the true underlying distribution $P(X|Y=\cmin)$ (unknown in practice).
Consequently, ROS does not change the $\mathbb{E}_D[d^{max}_\cmin]$ and the expected bias, albeit it enlarges the minority class size $|D_\cmin|$, i.e., $\mathbb{E}_D[|\zeta^*-\hat{\zeta}|] = \mathbb{E}_{D'}[|\zeta^*-\hat{\zeta}|]$.
Moreover, due to the duplication-based design, ROS can also cause the classifier to overfit the pattern of minority class samples~\cite{he2008overview,he2013overview}.

The Synthetic Minority Over-sampling TechniquE~\cite{chawla2002smote} (SMOTE) is one of the most popular methods to prevent the risk of overfitting induced by ROS.
It improves the na\"ive ROS by finding the k-nearest neighbors (kNNs) for all minority examples and performing linear interpolation between a seed example and one of its kNNs.
SMOTE has been proved to be effective in preventing overfitting and improving the quality of minority class representations in many real-world applications~\cite{haixiang2017learning-from-imb-review}.
However, from \figurename~\ref{fig:intercb-smote} we can observe that although the distribution of the over-sampled minority class is smoothed by SMOTE, the $\mathbb{E}_D[d^{max}_\cmin]$ still remains unchanged, so does the expected bias $\mathbb{E}_D[|\zeta^*-\hat{\zeta}|]$.
Therefore, SMOTE only partially solves the problems that exist in ROS.

\textbf{Hybrid-sampling.}
Finally, we consider a simple combination of RUS and ROS called random hybrid-sampling (RHS).
The results are shown in \figurename~\ref{fig:intercb-rhs}.
We can see that RHS is also an effective way to mitigate the decision bias (51\% reduction, 0.706$\to$0.346) as it incorporates RUS.
Compared with the strict under-sampling $|D_\cmaj|\xrightarrow[]{\text{RUS}}|D_\cmin|$, the average-targeted under-sampling in RHS is less aggressive, thus preventing the introduction of the opposite bias towards $\cmaj$ as in RUS.
As a result, RHS yields more stable decision boundaries over multiple runs, where the variance (0.035) of RHS is lower than that of RUS (0.045).
In terms of over-sampling, RHS adopts the na\"ive ROS, but produces fewer new samples to mitigate potential overfitting problems.
The reasons for not using more advanced over-sampling techniques (e.g., SMOTE~\cite{chawla2002smote} and its variants~\cite{he2008adasyn,han2005borderline-smote}) are: 
\textbf{(1)} as stated above, these techniques do not help to reduce decision bias, 
\textbf{(2)} they prerequisite a well-defined distance metric and introduce additional computational cost, and 
\textbf{(3)} with proper processing (i.e., the perturbation-based data augmentation described in \S~\ref{sec:pbda}), duplication-based ROS has the potential to increase $\mathbb{E}_D[d^{max}_\cmin]$ and thus reduces bias.

From the previous discussion, we can conclude that \textit{the key to mitigating decision bias is the ability to equalize $\mathbb{E}[d^{max}_\cmin]$ and $\mathbb{E}[d^{max}_\cmaj]$}.
This can be done by scaling down $\mathbb{E}[d^{max}_\cmaj]$ and/or scaling up $\mathbb{E}[d^{max}_\cmin]$.
The former can be achieved by RUS, but with significant information loss and variance.
The latter is a difficult problem as the underlying minority distribution is unknown in practice, existing methods like ROS and SMOTE fundamentally generate new examples within the existing ones, thus cannot change $\mathbb{E}[d^{max}_\cmin]$. 
RHS combines RUS and ROS and achieves more robust adjustification of $\mathbb{E}[d^{max}_\cmaj]$, but new techniques are still needed to increase $\mathbb{E}[d^{max}_\cmin]$.

\subsection{Perturbation-based Data Augmentation (PBDA).}
\label{sec:pbda}

\begin{figure*}[t]
  \centering
  \subfigure[Random Under-sampling.]{
    \includegraphics[width=\quadwidth]{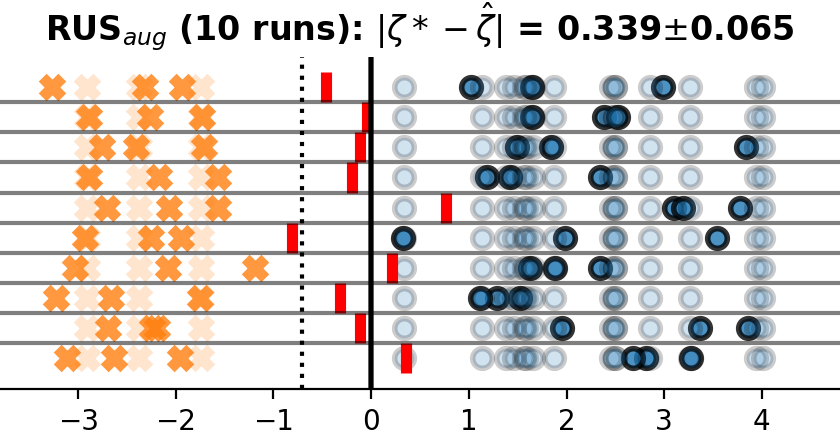}
    \label{fig:intercb-aug-rus}
  }
  \subfigure[Random Over-sampling.]{
    \includegraphics[width=\quadwidth]{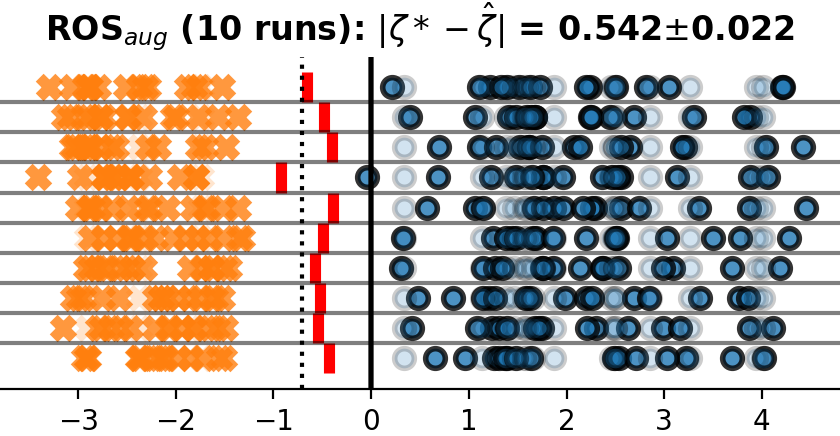}
    \label{fig:intercb-aug-ros}
  }
  \subfigure[SMOTE Over-sampling~\cite{chawla2002smote}.]{
    \includegraphics[width=\quadwidth]{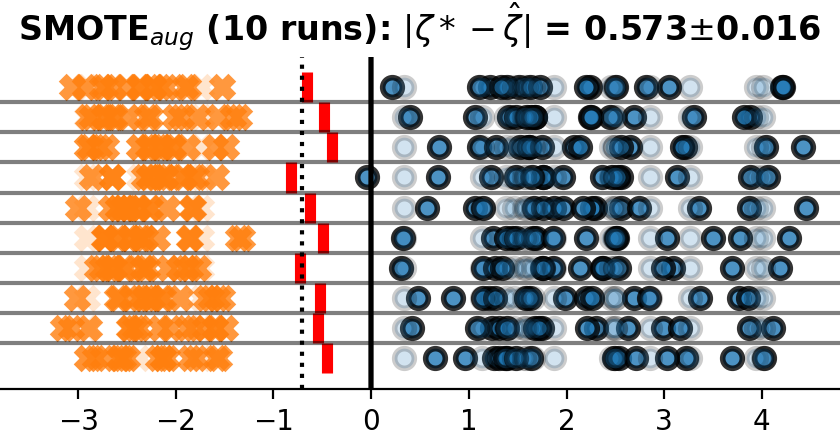}
    \label{fig:intercb-aug-smote}
  }
  \subfigure[Random Hybrid-sampling.]{
    \includegraphics[width=\quadwidth]{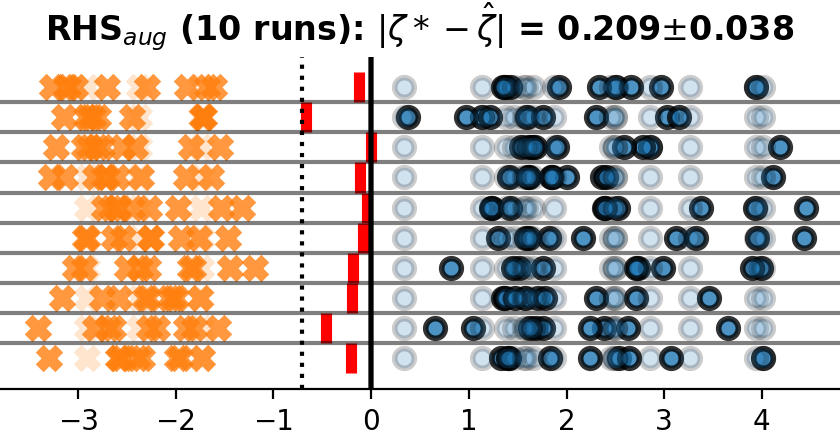}
    \label{fig:intercb-aug-rhs}
  }
  \caption{
  A comparison of different inter-class balancing strategies \textit{with data augmentation}. 
  Each subplot shows the results of the corresponding method in 10 independent runs.
  Same as in \figurename~\ref{fig:intercb}, optimal $\zeta^{*}$ (solid line), original $\hat{\zeta}_D$ (dotted line) and new $\hat{\zeta}_{D'_\text{aug}}$s (red lines) are also included.
  }
  \label{fig:intercb-aug}
\end{figure*}

Motivated by the above analysis, we propose a simple way to adjust $\mathbb{E}[d^{max}_\cmin]$, namely perturbation-based data augmentation (PBDA).
In this case, we add random Gaussian noises $x^p$ to the resampled dataset $D'$ and get augmented data $D'_\text{aug}$.
Formally, 
\begin{equation}
    D'_\text{aug}: \{(x+x^p, y)\}, x^p\sim\mathcal{N}(0,\sigma^p), (x, y)\in D',
\end{equation}
where $\sigma^p$ is a hyper-parameter that controls the intensity of perturbation.
But why does it help to increase $\mathbb{E}[d^{max}_\cmin]$?

\textbf{Why PBDA works?}
Recall that we define the support vector of minority class $s_+^\text{sup}: (x_+^\text{sup}, \cmin)$ and majority class $s_-^\text{sup}: (x_-^\text{sup}, \cmaj)$.
Without loss of generality, we can always assume that the optimal decision boundary is at the origin of the coordinate axis, i.e., $\zeta^*=0$ and $\mu_-+\mu_+=0$, so the bias $|\zeta^*-\hat{\zeta}|=|(x_+^\text{sup}+x_-^\text{sup})/2|, x_+^\text{sup}<0$.
Therefore, in order to reduce the bias, we want to increase $x_+^\text{sup}$, which cannot be done by directly using the over/hybrid-sampling methods discussed before.
However, note that after duplication-based over-sampling, there will be multiple replications of $s_+^\text{sup}$.
Suppose the number of replications is $n^\text{rep}$, after applying PBDA, these replications can be seen as $n^\text{rep}$ i.i.d. samples from $\mathcal{N}(x_+^\text{sup}, \sigma^p)$.
Therefore, the new support vector on the augmented data $x_+^\text{newsup} = max(\text{augmented replications of }x_+^\text{sup})$.
As proved in~\cite{kamath2015bounds}, $\mathbb{E}[x_+^\text{newsup}-x_+^\text{sup}]$ ($\mathbb{E}[\Delta x_+^\text{sup}]$) is bounded by:
\begin{equation}
    0 \leq
    \frac{1}{\sqrt{\pi\text{log}2}}\sigma^p\sqrt{\text{log}n^\text{rep}} \leq 
    \mathbb{E}[\Delta x_+^\text{sup}] \leq
    \sqrt{2}\sigma^p\sqrt{\text{log}n^\text{rep}}.
    \label{eq6}
\end{equation}
This demonstrates that \textit{duplication-based over-sampling with PBDA is effective in increasing $x_+^\text{sup}$, and thus yields a less biased decision boundary}.
Note that the expected bias reduction is bounded by terms related to the hyper-parameter $\sigma^p$ and the number of minority support vector replications $n^\text{rep}$.

We carry out further experiments to validate the above analysis, and the results are shown in \figurename~\ref{fig:intercb-aug}.
Specifically, we extend the experiments in \figurename~\ref{fig:intercb} by applying PBDA on the resampled data, the $\sigma^p$ is set to $0.2$.
First, it can be observed from \figurename~\ref{fig:intercb-aug-rus} that PBDA is not helpful when used in conjunction with RUS.
It is not a surprising result as $n^\text{rep}=1$ for RUS, substituting into Eq.~\eqref{eq6}, we have $\mathbb{E}[\Delta x_+^\text{sup}] \leq \sqrt{2}\sigma^p\sqrt{\text{log}1} = 0$, i.e., the PBDA does not reduce the expected bias when $n^\text{rep}=1$.
In addition, the performance of RUS becomes even more unstable (variance increases from 0.045 to 0.065) due to the additional uncertainty introduced by PBDA.
As for ROS with $n^\text{rep}>1$, PBDA significantly reduces its decision bias by 0.164 (0.706$\to$0.542, \figurename~\ref{fig:intercb-aug-ros}).
Similarly, the bias of SMOTE is also reduced but by a smaller margin (0.133) due to the smaller $n^\text{rep}$ compared with ROS (\figurename~\ref{fig:intercb-aug-smote}).
Finally, we find that RHS is the best performer cooperated with PBDA (\figurename~\ref{fig:intercb-aug-rhs}). 
It yields the smallest bias (0.209 $\pm$ 0.038) since it works in both directions: with RUS to adjust $\mathbb{E}[d^{max}_\cmaj]$, and with ROS+PBDA to adjust $\mathbb{E}[d^{max}_\cmin]$.

We note that aside of the classification errors of each base classifier, the diversity across the ensemble members is also a key factor that affects the performance of ensemble models~\cite{galar2012ensemble,zhang2012ensemble}.
Therefore, beyond mitigating decision bias, PBDA also plays an important role in diversifying the base classifiers of {\sc DuBE}.
It also serves as a technique to prevent the risk of overfitting introduced by duplication-based over-sampling.

\textbf{Formalization.}
In practice, the conditional underlying distribution $P(X|Y)$ is likely to vary in different classes, thus sample perturbation signals $x^p$ from a single Gaussian distribution $\mathcal{N}(0,\sigma^p)$ is insufficient.
In response to this, we add different perturbations for each class independently in {\sc DuBE}, the intensity is controlled by a single hyper-parameter $\alpha$:
\begin{equation}
    D'_\text{aug}: \{(x+\alpha \cdot x^p,c_i)\}, x^p\sim\mathcal{N}(0,\Sigma_{c_i}), \forall c_i \in \mathcal{Y},
    \label{eq7}
\end{equation}
where the $\Sigma_{c_i}$ is the covariance matrix estimated from $D_{c_i}$:
\begin{equation}
    \Sigma_{c_i}=\frac{1}{n_{i}-1} \sum_{j=1}^{n_{i}}\left(x_{j}-\mu_{i}\right)\left(x_{j}-\mu_{i}\right)^\top.
    \label{eq8}
\end{equation}
In Eq.~\eqref{eq8}, $n_{i}=|D_{c_i}|$ and $\mu_i$ is the centroid of the class $c_i$, i.e., $\mu_{i}=\sum_{j=1}^{n_{i}}\left(x_{j}\right)/n_{i}, x_j \in D_{c_i}$.
By doing so, we calibrate the perturbation using statistics of the base classes.

\subsection{Intra-class Balancing (IntraCB).}
\label{sec:intracb}

We now discuss the influence of different IntraCB strategies in \method.
Intra-class balancing in \method is achieved by assigning different sampling probability $w$ to each instance $(x, y)$ based on the prediction error w.r.t. a classifier $f$, i.e., an IntraCB strategy corresponds to a function $g(\cdot): \mathcal{X}\times\mathcal{Y}\times\mathcal{F} \to \mathbb{R}$.
Here the $\mathcal{F}$ is the hypothesis (model) space.

\textbf{Hard example mining.}
Let's first introduce the hard example mining (HEM) considered in \method.
Formally, for an input $x$ with label space $\mathcal{Y}:\{c_1, c_2, \cdots, c_m\}$, given a classifier $f(\cdot)$, the estimated probabilities are $f(x)=\textbf{p}=[p_1, p_2, ..., p_m]^\top$, where the $i$-th component $p_i \in [0, 1]$ is the estimated $P(y=c_i|x)$.
For a label $y=c_i$, we define the one-hot binarized label vector ${\bf y}=[y_1, y_2, ..., y_m]^\top$, where $y_i=1$ and other components are $0$.
Then the prediction error is defined as $\sum_{i=0}^{m} |p_i - y_i|$.
For simplicity, we use {\bf err}$(f(x), y)$ to denote the error of $(x,y)$ w.r.t. $f$.
Then in the $t$-th iteration of {\sc DuBE}, given the current ensemble classifier $F_{t-1}(\cdot)$, hard example mining corresponds to the following weighting function:
\begin{equation}
    g_\text{HEM}(x, y, F_{t-1}) = \textbf{err}(F_{t-1}(x), y),
\end{equation}
i.e., hard examples with larger prediction errors are more likely to be sampled and used in subsequent learning.

When the dataset contains no noise (i.e., has better separability), we show that with duplication-based over-sampling and PBDA, HEM is effective in term of further reducing decision bias.
As shown in \figurename~\ref{fig:intracb-clean}, the minority support vector $x_+^\text{sup}$ has the largest error among all minority class samples as it locates nearest to the decision boundary.
Suppose that we want to expand the minority class size from $n_+$ to $n'_+$, where $n'_+>n_+$.
With ROS, the expected number of $x_+^\text{sup}$ replications is $\mathbb{E}_\text{ROS}[n^\text{rep}]=n'_+/n_+$.
With ROS+HEM, this expectation becomes $\mathbb{E}_\text{ROS+HEM}[n^\text{rep}]=(e^\text{sup}_+ \cdot n'_+)/(\sum_{s_+\in D_\cmin}e_+)$, where $e$ represents the error of a sample $s$.
It is obvious that $\mathbb{E}_\text{ROS+HEM}[n^\text{rep}]>\mathbb{E}_\text{ROS}[n^\text{rep}]$ since $e^\text{sup}_+ \geq e_+, \forall s_+\in D_\cmin$.
Then according to Eq.~\eqref{eq6}, we can expect $\mathbb{E}_\text{ROS+HEM}[x_+^\text{sup}]>\mathbb{E}_\text{ROS}[x_+^\text{sup}]$ after PBDA, and therefore yields smaller bias.

Nevertheless, HEM can wrongly reinforce noise/outliers when working on a noisy dataset.
In \figurename~\ref{fig:intracb-noisy}, we modify the data in \figurename~\ref{fig:intracb-clean} by adding minority class noise samples.
It can be observed that these few noise samples contribute most of the prediction errors of the minority class.
Consequently, these outliers will dominate the resampling process with HEM.
In this case, the outliers will be duplicated multiple times, while other samples including the true support vector $x_+^\text{sup}$ will be ignored by HEM.
This is clearly not a good strategy as \textit{the extracted data will contain mainly noisy instances, which will greatly interfere with the subsequent learning process}.

\newcommand{\singlewidth}{0.97\linewidth}

\begin{figure}[t]
  \centering
  \subfigure[Without data noise.]{
    \includegraphics[width=\singlewidth]{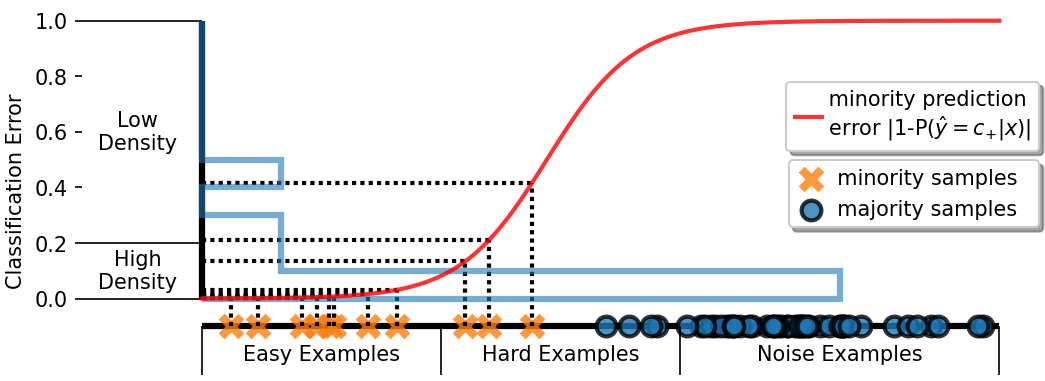}
  \label{fig:intracb-clean}
  }
  \subfigure[With data noise.]{
    \includegraphics[width=\singlewidth]{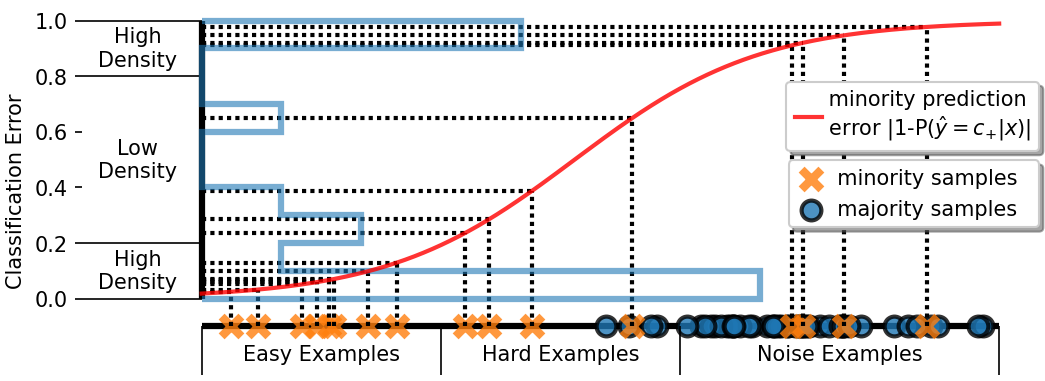}
  \label{fig:intracb-noisy}
  }
  \caption{
  The classification error (y-axis) distributions with/without presence of minority class noise.
  The prediction errors (probabilities) are obtained from a Logistic Regression classifier trained on the corresponding data.
  }
  \label{fig:intracb}
\end{figure}

\textbf{Soft hard example mining.}
However, we find that this can be prevented by considering the error density distribution.
After taking a closer look at the distribution of classification errors (y-axis) in \figurename~\ref{fig:intracb}, we have some interesting findings:
\textbf{(1)} 
hard examples are likely to be sparsely distributed in the region near the decision boundary (e.g., the "hard examples" area on the x-axis), where the estimated probability $P(\hat{Y}|X)$ also changes drastically.
All these lead to an even more sparse error distribution of hard examples (e.g., the "low density" area on the y-axis);
\textbf{(2)}
By comparison, both easy examples and noise examples are densely distributed in terms of classification error (e.g., the "high density" areas on the y-axis).
This is mainly because $P(\hat{Y}|X)$ varies only a little in both the "easy examples" and "noisy examples" regions as the classifier has a high confidence in the predictions of these samples.

Motivated by the above analysis, we propose to perform robust HEM by simply \textit{inversing the classification error density}.
By doing so, the easy and noisy examples with high error densities will be down-weighted.
Intuitively, this will prevent over-weighting of noisy examples while still emphasizing the importance of mining hard examples.
There are many ways to estimate the error distribution $E$, and herein we use a histogram to approximate $E$ for its simplicity and efficiency.
Formally, consider a histogram with $b$ bins, then the approximated error distribution is given by a vector $\widehat{E}_D=[\widehat{E}^1_D, \widehat{E}^2_D, ..., \widehat{E}^b_D] \in \mathbb{R}^b$, where the $i$-th component $\widehat{E}^i_D$ denotes the proportion of the number of samples in the $i$-th bin ($B^i$) to the total, i.e.,
\begin{equation}
    \widehat{E}^i_D = \frac{|B^i|}{|D|}; B^i = \{s_j | \frac{i-1}{b} \leq e_j < \frac{i}{b}, \forall s_j \in D\}.
    \label{eq10}
\end{equation}
Here the $e_j$ is the error of $s_j$, i.e., $e_j=\textbf{err}(F_{t-1}(x_j), y_j)$.
With Eq.~\eqref{eq10}, we can now formally define the Soft Hard Example Mining (SHEM) in DuBE: 
\begin{equation}
    g_\text{SHEM}(x, y, F_{t-1}) = 1 / \widehat{E}_D^{\lceil \textbf{err}(F_{t-1}(x), y)/b \rceil}.
    \label{eq11}
\end{equation}
The $\lceil \textbf{err}(F_{t-1}(x), y)/b \rceil$ in Eq.~\eqref{eq11} indicates the bin that $(x, y)$ belongs to, i.e., instances are assigned sampling weights corresponding to the inverse of their classification error density.
To this point, we can summarize the \method in Alg.~\ref{alg:dube}.
The implementation\footnote{\color{blue}\href{\codeurl}{\underline{\tt \codeurl}}} and documentation\footnote{\color{blue}\href{\docurl}{\underline{\tt \docurl}}} are open-sourced.
Please refer to \S~\ref{sec:ablation} for more related results and discussions.

\begin{algorithm}[h]
    \caption{{\tt Duple-Balanced Ensemble Training}}
    \label{alg:dube}
    \begin{algorithmic}[1]
        \Require \textrm{training set} $D$, \textrm{ensemble size} $k$, \textrm{perturbation hyper-parameter} $\alpha$, \textrm{number of bins in the error histogram} $b$\;
        \State Derive covariance matrices $\Sigma_{c_1}, \Sigma_{c_2}, ..., \Sigma_{c_m}$. \Comment Eq.~\eqref{eq8}\;
        \State Train the first base classifier $f_1(\cdot)$ on $D$\;
        \For{t=2 to $k$}
            \State Update current ensemble $F_{t-1}(x) = \frac{1}{t-1}\sum_{i=1}^{t-1} f_i(x)$.\;
            \State \textit{\# Inter-class Balancing }$_\text{(RUS/ROS/RHS)}$ \Comment Section \ref{sec:intercb}\;
            \State Set the target class size $n$ for resampling.\;
            \State \textit{\# Intra-class Balancing }$_\text{(HEM/SHEM)}$ \Comment Section \ref{sec:intracb}\;
            \State Set the instance-wise weight $w$ w.r.t. $F_{t-1}(\cdot)$.\;
            \State \textit{\# Duple-Balanced Resampling}\;
            \State Initialization: $D' \Leftarrow \emptyset$\;
            \For{i=1 to $m$}
                \State $D'_{c_i} \Leftarrow n$ instances sampled from $D_{c_i}$ w.r.t. $w$.\;
                \State \textit{\# Data Augmentation} \Comment Section \ref{sec:pbda}\;
                \State $D'_{c_i,\text{aug}} \Leftarrow \{(x+\alpha \cdot x^p,c_i)\}_{x^p\sim\mathcal{N}(0,\Sigma_{c_i})}$ \Comment Eq.~\eqref{eq7}\;
                \State $D' \Leftarrow D' \cup D'_{c_i,\text{aug}}$\;
            \EndFor
            \State Train a new base classifier $f_t(\cdot)$ with $D'$.\;
        \EndFor
        \State \Return {Ensemble classifer $F_k(\cdot)$}
    \end{algorithmic}
\end{algorithm}
\begin{figure*}[t]
  \centering
  \includegraphics[width=\linewidth]{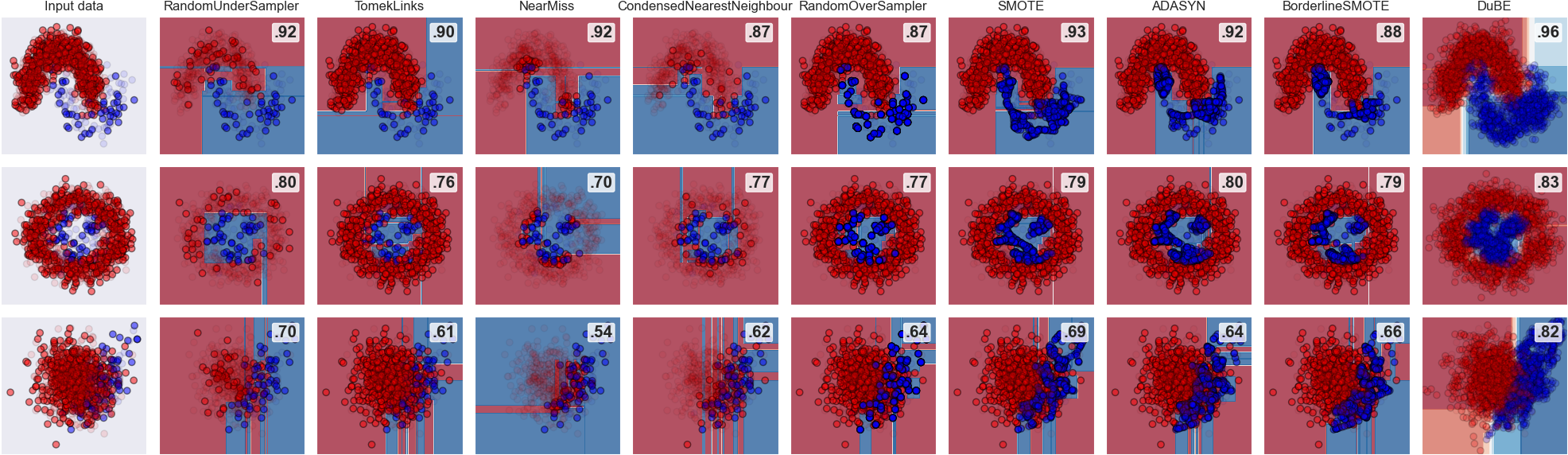}
  \caption{
  Comparisons of \method with 8 representative resampling methods (under-sampling: {\sc RUS}, {\sc TomekLink}~\cite{tomek1976tomeklink}, {\sc NearMiss}~\cite{mani2003nearmiss}, {\sc Condense}~\cite{hart1968cnn}, over-sampling: {\sc ROS}, {\sc Smote}~\cite{chawla2002smote}, {\sc Adasyn}~\cite{he2008adasyn}, and {\sc BorderSmote}~\cite{han2005borderline-smote}) on 3 synthetic imbalanced datasets with different level of underlying class distribution overlapping (less/mid/highly overlapped in 1st/2nd/3rd row).
  The opaque dots are the (resampled) training set and the translucent dots represent the test set.
  The number in the upper right of each subfigure is the test macro AUROC obtained by the classifier. 
  Best viewed in color.
  }
  \label{fig:comp-resamp}
\end{figure*}

\section{Experiments}
\label{sec:experiments}

In this section, we conduct experiments on both synthetic and real-world tasks to answer following research questions:
\begin{itemize}
    \item \textbf{\textit{RQ1:}} 
    How does DuBE's behavior different from existing IL techniques?
    ($\triangleright$ Section~\ref{sec:exp-synthetic})
    \item \textbf{\textit{RQ2:}} 
    Is DuBE effective in real-world tasks?($\triangleright$ \S~\ref{sec:exp-real})
    \item \textbf{\textit{RQ3:}} 
    Does DuBE show performance/efficiency advantages over its resampling-based ($\triangleright$ Section~\ref{sec:exp-real-resampling}) and ensemble-based ($\triangleright$ Section~\ref{sec:exp-real-ensemble}) counterparts?
    \item \textbf{\textit{RQ4:}} 
    How do different choices of inter- and intra-class balancing strategies and hyperparameters affect the performance of DuBE?
    ($\triangleright$ Section~\ref{sec:ablation})
\end{itemize}


\begin{table*}[t]
  \centering
  \caption{Characteristics of the real-world class-imbalanced datasets.}
  \label{tab:datasets}
  \begin{tabular}{c|c|c|c|c|c|c|c}
  \toprule
  \textbf{Dataset} & \textbf{\#Samples} & \textbf{\#Features} & \textbf{Class Distribution} & \textbf{Imbalance Ratio} & \textbf{Task} & \textbf{Domain} \\ \midrule
  {\it ecoli}         & 336    & 7   & {\tt 301/35}      & {\tt 8.60/1.00}      & target: imU        & Bioinformatics. \\
  {\it pen-digits}    & 10,992 & 16  & {\tt 9,937/1,055} & {\tt 9.42/1.00}      & target: 5          & Computer. \\
  {\it spectrometer}  & 531    & 93  & {\tt 486/45}      & {\tt 10.80/1.00}     & target: >=44       & Astronomy. \\
  {\it scene}         & 2,407  & 284 & {\tt 2,230/177}   & {\tt 12.60/1.00}     & target: >one label & Geography. \\
  {\it libras-move}   & 360    & 90  & {\tt 336/24}      & {\tt 14.00/1.00}     & target: 1          & Physics. \\
  {\it oil}           & 937    & 49  & {\tt 896/41}      & {\tt 21.85/1.00}     & target: raw        & Environment. \\
  {\it letter-img}    & 20,000 & 16  & {\tt 19,266/734}  & {\tt 26.25/1.00}     & target: Z          & Computer. \\
  {\it ozone-level}   & 2,536  & 72  & {\tt 2,463/73}    & {\tt 33.74/1.00}     & target: raw        & Climatology. \\
  {\it balance-scale} & 625    & 4   & {\tt 288/288/49}  & {\tt 5.88/5.88/1.00} & target: raw        & Psychology. \\
  {\it cmc}           & 1,473  & 24  & {\tt 629/511/333} & {\tt 1.89/1.53/1.00} & target: raw        & Sociology. \\
  \bottomrule
  \end{tabular}
\end{table*}

\begin{table*}[h]
\centering
\scriptsize
\caption{
    Comparisons of DuBE with representative resampling-based IL solutions.
    The macro-AUROC (mean$\pm$std), the number of training samples, and the time used for resampling are reported.
    The best results (among baselines) are marked in \textbf{bold} (\underline{underlined}).
}
\label{tab:results-resamp}
\begin{tabular}{c|c|ccccc|c|c}
\toprule
\multirow{2}{*}{\textbf{Category}} & \multirow{2}{*}{\textbf{Method}}                 & \multicolumn{5}{c|}{\textbf{Base Learning Algorithm}}                                                                                                                       & \textbf{\#Training} & \textbf{Resampling} \\ \cline{3-7} 
                                    &                                                  & \multicolumn{1}{c|}{\textbf{MLP}}    & \multicolumn{1}{c|}{\textbf{KNN}}    & \multicolumn{1}{c|}{\textbf{DT}}     & \multicolumn{1}{c|}{\textbf{BST}}    & \textbf{BAG}    & \textbf{Samples}    & \textbf{Time (ms)}  \\ \midrule
No-resampling & - & \multicolumn{1}{c|}{0.499$\pm$0.004} & \multicolumn{1}{c|}{0.628$\pm$0.006} & \multicolumn{1}{c|}{0.669$\pm$0.012} & \multicolumn{1}{c|}{0.694$\pm$0.013} & 0.684$\pm$0.017 & 2029 & - \\ \hline
\multirow{4}{*}{Under-sampling}    & {\sc Rus}                                        & \multicolumn{1}{c|}{0.501$\pm$0.005} & \multicolumn{1}{c|}{0.660$\pm$0.006} & \multicolumn{1}{c|}{0.674$\pm$0.032} & \multicolumn{1}{c|}{\uline{0.702$\pm$0.030}} & 0.683$\pm$0.030 & 116                 & 0.72                \\ 
                                    & {\sc TomekLink}~\cite{tomek1976tomeklink}        & \multicolumn{1}{c|}{0.509$\pm$0.012} & \multicolumn{1}{c|}{0.506$\pm$0.000} & \multicolumn{1}{c|}{0.569$\pm$0.006} & \multicolumn{1}{c|}{0.566$\pm$0.006} & 0.541$\pm$0.020 & 2,008               & 62.28               \\ 
                                    & {\sc NearMiss}~\cite{mani2003nearmiss}           & \multicolumn{1}{c|}{0.505$\pm$0.009} & \multicolumn{1}{c|}{0.444$\pm$0.000} & \multicolumn{1}{c|}{0.472$\pm$0.011} & \multicolumn{1}{c|}{0.489$\pm$0.013} & 0.487$\pm$0.011 & 116                 & 4.65                \\ 
                                    & {\sc Condense}~\cite{hart1968cnn}                & \multicolumn{1}{c|}{0.501$\pm$0.002} & \multicolumn{1}{c|}{0.517$\pm$0.000} & \multicolumn{1}{c|}{0.655$\pm$0.008} & \multicolumn{1}{c|}{0.660$\pm$0.015} & 0.655$\pm$0.008 & 311                 & 12157.64            \\ 
                                    \hline
\multirow{4}{*}{Over-sampling}     & {\sc Ros}                                        & \multicolumn{1}{c|}{0.504$\pm$0.003} & \multicolumn{1}{c|}{0.571$\pm$0.000} & \multicolumn{1}{c|}{0.610$\pm$0.010} & \multicolumn{1}{c|}{0.609$\pm$0.020} & 0.595$\pm$0.007 & 3,942               & 1.64                \\ 
                                    & {\sc Smote}~\cite{chawla2002smote}               & \multicolumn{1}{c|}{0.498$\pm$0.004} & \multicolumn{1}{c|}{0.637$\pm$0.013} & \multicolumn{1}{c|}{\uline{0.680$\pm$0.013}} & \multicolumn{1}{c|}{0.629$\pm$0.002} & 0.629$\pm$0.028 & 3,942               & 3.99                \\ 
                                    & {\sc Adasyn}~\cite{he2008adasyn}                 & \multicolumn{1}{c|}{0.515$\pm$0.030} & \multicolumn{1}{c|}{0.630$\pm$0.003} & \multicolumn{1}{c|}{0.618$\pm$0.025} & \multicolumn{1}{c|}{0.639$\pm$0.018} & 0.636$\pm$0.017 & 3,942               & 6.01                \\ 
                                    & {\sc BorderSmote}~\cite{han2005borderline-smote} & \multicolumn{1}{c|}{0.511$\pm$0.017} & \multicolumn{1}{c|}{0.592$\pm$0.004} & \multicolumn{1}{c|}{0.592$\pm$0.009} & \multicolumn{1}{c|}{0.579$\pm$0.016} & 0.599$\pm$0.038 & 3,942               & 6.20                \\ 
                                    \hline
Over-sampling                      & {\sc SmoteEnn}~\cite{batista2004smoteenn}        & \multicolumn{1}{c|}{0.514$\pm$0.025} & \multicolumn{1}{c|}{\uline{0.670$\pm$0.009}} & \multicolumn{1}{c|}{0.628$\pm$0.020} & \multicolumn{1}{c|}{0.619$\pm$0.023} & \uline{0.693$\pm$0.017} & 3,500               & 270.76              \\ 
+ Cleaning                         & {\sc SmoteTomek}~\cite{batista2003smotetomek}    & \multicolumn{1}{c|}{\uline{0.518$\pm$0.028}} & \multicolumn{1}{c|}{0.618$\pm$0.009} & \multicolumn{1}{c|}{0.625$\pm$0.017} & \multicolumn{1}{c|}{0.607$\pm$0.016} & 0.641$\pm$0.021 & 3,939               & 257.40              \\ 
                                    \hline
\multirow{3}{*}{Ours}              & {\sc Dube}$_{10, \text{RUS}}$                     & \multicolumn{1}{c|}{0.610$\pm$0.076} & \multicolumn{1}{c|}{0.723$\pm$0.005} & \multicolumn{1}{c|}{0.794$\pm$0.014} & \multicolumn{1}{c|}{\bf 0.800$\pm$0.005} & 0.804$\pm$0.006 & 116$\times$10       & 1.39                \\ 
                                    & {\sc Dube}$_{10, \text{ROS}}$                    & \multicolumn{1}{c|}{0.534$\pm$0.022} & \multicolumn{1}{c|}{0.697$\pm$0.025} & \multicolumn{1}{c|}{0.797$\pm$0.026} & \multicolumn{1}{c|}{0.792$\pm$0.026} & {\bf 0.812$\pm$0.005} & 3,942$\times$10     & 1.69                \\ 
                                    & {\sc Dube}$_{10, \text{RHS}}$                    & \multicolumn{1}{c|}{\bf 0.659$\pm$0.018} & \multicolumn{1}{c|}{\bf 0.724$\pm$0.009} & \multicolumn{1}{c|}{\bf 0.801$\pm$0.009} & \multicolumn{1}{c|}{0.798$\pm$0.016} & 0.802$\pm$0.007 & 2,029$\times$10     & 1.52                \\ 
                                    \bottomrule
\end{tabular}
\end{table*}

\newcommand{\resultwidthvis}{0.195\linewidth}

\subsection{Experiment on Synthetic Datasets}
\label{sec:exp-synthetic}

{\bf Setup details.} 
To intuitively demonstrate how different IL methods work in IL problems, we first show some visualizations on a series of synthetic datasets.
We create three 2-dimensional classification tasks with 1,100 samples from 2 classes, the class distribution is {\tt 1000/100}, i.e., imbalance ratio (IR) = 10.
Note that there exists class overlapping in the synthetic dataset, as shown in Figure \ref{fig:comp-resamp}.
In this experiment, \method is compared with 8 representative resampling methods, 4 of which are under-sampling (i.e., {\sc RUS}, {\sc TomekLink}~\cite{tomek1976tomeklink}, {\sc NearMiss}~\cite{mani2003nearmiss}, {\sc Condense}~\cite{hart1968cnn}) and the other 4 are over-sampling (i.e., {\sc ROS}, the widely used {\sc Smote}~\cite{chawla2002smote}, and its variants {\sc Adasyn}~\cite{he2008adasyn} and {\sc BorderSmote}~\cite{han2005borderline-smote}).
All methods included in this experiment are deployed with decision trees as base classifiers.
The ensemble size of \method is 5.
We perform 50\%/50\% train/test split on each dataset, and use translucent dots to represents the test set in \figurename~\ref{fig:comp-resamp}.
For each IL algorithm, we plot the resampled training dataset, the learned decision boundary, and the macro-averaged AUROC (a class-balanced metric) score on the test set in \figurename~\ref{fig:comp-resamp}.

{\bf Visualization \& analysis.}
As mentioned earlier, RUS randomly removes the majority class sample, which helps mitigate decision bias but leads to severe information loss.
This may cause the classifier to overfit the selected subset and generate unstable learning boundary.
{\sc TomekLink}~\cite{tomek1976tomeklink} performs under-sampling by detecting "TomekLinks", which exists if two samples of different class are the nearest neighbors of each other, i.e., it removes the majority samples that locate in the overlapping area.
But we can see that only a few majority samples are discarded, which does not help with mitigating the decision bias. 
It is therefore outperformed by RUS, especially on highly-overlapped datasets (e.g., 3rd row).
In contrast to {\sc TomekLink}, both {\sc NearMiss}~\cite{mani2003nearmiss} and {\sc Condense}~\cite{hart1968cnn} aim to select the majority samples that are closest to the minority ones.
They assume these samples are likely to be support vectors that help classification.
However, this assumption clearly does not hold when the data sets overlap: both methods end up dropping most of the majority samples that reflect the original distribution, which greatly interferes with the learning process.

Compared with under-sampling, over-sampling methods typically produce more stable decision bounds as they only add new samples and keep the original dataset untouched.
ROS works by simply replicating existing minority instances, which may cause the learner over-fit the pattern of minority class samples, as shown in the [1st row, RandomOverSampler] sub-figure.
Advanced over-sampling techniques prevent this by performing neighborhood-based interpolation instead of duplication.
{\sc Smote}~\cite{chawla2002smote} works by repeatedly selecting seed instances in the minority class, and then generating synthetic samples on the connection line between the seed and one of its nearest neighbors.
This effectively smooths out the minority distribution after over-sampling.
{\sc Adasyn}~\cite{he2008adasyn} and {\sc BorderSmote}~\cite{han2005borderline-smote} further improve {\sc Smote} by focusing on borderline minority examples for interpolation.
However, such strategy may result in generating massive instances near minority outliers (e.g., data points at the upper left end of the "$\cup$"-shaped distribution of the 1st dataset).

Finally, we can observe that \method ($k=5$, with RHS, SHEM and PBDA) achieves the best performance in all three tasks.
Compared with strict under-sampling, the hybrid-sampling essentially maintains the original distribution structure of the data.
On the other hand, compared with {\sc Adasyn} and {\sc BorderSmote}, the usage of SHEM prevents \method from generating synthetic instances near minority outliers.
Moreover, PBDA improves the coverage of minority class in the feature space, thus alleviating decision bias. 
It also enables data-level regularization and stabilizes the decision boundary of {\sc DuBE}.

\subsection{Experiment on Real-world Datasets}
\label{sec:exp-real}

\begin{table*}[t]
  \centering
  \scriptsize
  \caption{
      Comparisons of DuBE with representative ensemble IL solutions.
      For each dataset, we report the generalized F1-score, MCC, and macro-AUROC scores (mean$\pm$std) in the 1st/2nd/3rd row.
      The best and second best results are marked in \textbf{bold} and \uline{underlined}.
  }
  \label{tab:results-ens}
  \begin{tabular}{cc|cccccccc|c|c}
  \toprule
  \multicolumn{2}{c|}{\diagbox[innerleftsep=1pt,width=8.5em,height=2.5em]{\bf Task$\times$Metric}{\bf Method}}
  & {\sc RusBst}  & {\sc OverBst} & {\sc SmoteBst} & {\sc UnderBag} & {\sc OverBag} & {\sc SmoteBag} & {\sc Cascade} & {\sc Spe}     & {\sc DuBE}  & {\bf $\Delta_\text{mean}$}  \\
  \midrule
  \multirow{3}{*}{\textit{ecoli $\times$}}        & F1-score          & .712$\pm$.013 & .736$\pm$.024 & .705$\pm$.020  & \uline{.772$\pm$.017}  & .754$\pm$.014 & .764$\pm$.012  & .715$\pm$.015 & .745$\pm$.023 & \textbf{.778$\pm$.004} &   5.48\% \\
                                                 & MCC         & .437$\pm$.026 & .472$\pm$.047 & .413$\pm$.039  & \uline{.571$\pm$.026}  & .518$\pm$.023 & .530$\pm$.026  & .508$\pm$.023 & .508$\pm$.046 & \textbf{.578$\pm$.010} &  17.92\% \\
                                                 & AUROC       & .756$\pm$.015 & .735$\pm$.020 & .716$\pm$.018  & .851$\pm$.007  & .727$\pm$.022 & .758$\pm$.004  & \uline{.858$\pm$.013} & .804$\pm$.026 & \textbf{.882$\pm$.008} &  14.21\% \\ \hline
  \multicolumn{2}{c|}{\multirow{3}{*}{\textit{pen-digits}}}    & .938$\pm$.009 & .978$\pm$.002 & .976$\pm$.002  & .980$\pm$.001  & .982$\pm$.001 & .983$\pm$.001  & .986$\pm$.001 & \uline{.989$\pm$.001} & \textbf{.996$\pm$.000} &   1.98\% \\
  \multicolumn{2}{c|}{}                                        & .876$\pm$.017 & .956$\pm$.003 & .952$\pm$.003  & .959$\pm$.002  & .964$\pm$.002 & .966$\pm$.002  & .973$\pm$.002 & \uline{.979$\pm$.001} & \textbf{.991$\pm$.001} &   4.09\% \\
  \multicolumn{2}{c|}{}                                        & .953$\pm$.004 & .973$\pm$.001 & .978$\pm$.001  & .983$\pm$.001  & .972$\pm$.001 & .975$\pm$.001  & \uline{.988$\pm$.001} & .986$\pm$.002 & \textbf{.995$\pm$.000} &   1.98\% \\ \hline
  \multicolumn{2}{c|}{\multirow{3}{*}{\textit{spectrometer}}}  & .800$\pm$.048 & .783$\pm$.028 & .770$\pm$.026  & .762$\pm$.003  & .786$\pm$.008 & .801$\pm$.045  & \uline{.901$\pm$.025} & .880$\pm$.016 & \textbf{.915$\pm$.013} &  13.30\% \\
  \multicolumn{2}{c|}{}                                        & .605$\pm$.097 & .568$\pm$.055 & .541$\pm$.052  & .553$\pm$.008  & .604$\pm$.025 & .613$\pm$.087  & \uline{.804$\pm$.048} & .762$\pm$.032 & \textbf{.831$\pm$.026} &  34.11\% \\
  \multicolumn{2}{c|}{}                                        & .835$\pm$.056 & .767$\pm$.029 & .787$\pm$.031  & .850$\pm$.011  & .729$\pm$.002 & .762$\pm$.048  & \uline{.919$\pm$.016} & .907$\pm$.013 & \textbf{.927$\pm$.015} &  13.87\% \\ \hline
  \multicolumn{2}{c|}{\multirow{3}{*}{\textit{scene}}}         & .535$\pm$.019 & \uline{.576$\pm$.000} & .554$\pm$.003  & \uline{.576$\pm$.010}  & .549$\pm$.006 & .573$\pm$.010  & .537$\pm$.006 & .546$\pm$.006 & \textbf{.601$\pm$.011} &   8.16\% \\
  \multicolumn{2}{c|}{}                                        & .114$\pm$.031 & .161$\pm$.005 & .112$\pm$.009  & \uline{.200$\pm$.022}  & .170$\pm$.004 & .196$\pm$.017  & .166$\pm$.013 & .199$\pm$.014 & \textbf{.204$\pm$.024} &  29.64\% \\
  \multicolumn{2}{c|}{}                                        & .585$\pm$.020 & .564$\pm$.003 & .565$\pm$.008  & .649$\pm$.016  & .537$\pm$.004 & .553$\pm$.007  & .640$\pm$.010 & \uline{.670$\pm$.013} & \textbf{.678$\pm$.014} &  14.59\% \\ \hline
  \multicolumn{2}{c|}{\multirow{3}{*}{\textit{libras-move}}}   & .727$\pm$.040 & \uline{.847$\pm$.017} & .827$\pm$.022  & .791$\pm$.055  & .827$\pm$.028 & .801$\pm$.023  & .791$\pm$.036 & .820$\pm$.041 & \textbf{.930$\pm$.014} &  15.87\% \\
  \multicolumn{2}{c|}{}                                        & .470$\pm$.069 & \uline{.697$\pm$.035} & .654$\pm$.044  & .600$\pm$.100  & .684$\pm$.047 & .643$\pm$.038  & .610$\pm$.069 & .649$\pm$.079 & \textbf{.860$\pm$.027} &  39.19\% \\
  \multicolumn{2}{c|}{}                                        & .751$\pm$.026 & .820$\pm$.013 & .822$\pm$.022  & .862$\pm$.034  & .762$\pm$.032 & .735$\pm$.024  & \uline{.886$\pm$.039} & .869$\pm$.035 & \textbf{.933$\pm$.021} &  15.16\% \\ \hline
  \multicolumn{2}{c|}{\multirow{3}{*}{\textit{oil}}}           & .578$\pm$.033 & .611$\pm$.023 & .683$\pm$.018  & .597$\pm$.030  & .640$\pm$.024 & \uline{.702$\pm$.049}  & .576$\pm$.014 & .607$\pm$.013 & \textbf{.707$\pm$.006} &  13.86\% \\
  \multicolumn{2}{c|}{}                                        & .224$\pm$.073 & .221$\pm$.046 & .370$\pm$.038  & .304$\pm$.045  & .368$\pm$.039 & \textbf{.434$\pm$.089}  & .286$\pm$.018 & .319$\pm$.020 & \uline{.420$\pm$.013} &  39.82\% \\
  \multicolumn{2}{c|}{}                                        & .694$\pm$.062 & .615$\pm$.025 & .705$\pm$.028  & .784$\pm$.027  & .592$\pm$.018 & .655$\pm$.044  & .785$\pm$.017 & \uline{.793$\pm$.013} & \textbf{.833$\pm$.012} &  19.89\% \\ \hline
  \multicolumn{2}{c|}{\multirow{3}{*}{\textit{letter-img}}}    & .826$\pm$.044 & .942$\pm$.002 & .940$\pm$.003  & .879$\pm$.008  & .949$\pm$.000 & .960$\pm$.004  & .943$\pm$.001 & \uline{.972$\pm$.001} & \textbf{.979$\pm$.003} &   5.99\% \\
  \multicolumn{2}{c|}{}                                        & .667$\pm$.072 & .885$\pm$.003 & .879$\pm$.005  & .776$\pm$.013  & .901$\pm$.001 & .922$\pm$.007  & .889$\pm$.001 & \uline{.945$\pm$.003} & \textbf{.959$\pm$.005} &  13.05\% \\
  \multicolumn{2}{c|}{}                                        & .896$\pm$.023 & .932$\pm$.001 & .947$\pm$.002  & .969$\pm$.002  & .917$\pm$.001 & .938$\pm$.005  & \uline{.978$\pm$.002} & .973$\pm$.003 & \textbf{.987$\pm$.002} &   4.64\% \\ \hline
  \multicolumn{2}{c|}{\multirow{3}{*}{\textit{ozone-level}}}   & .546$\pm$.014 & .593$\pm$.004 & .592$\pm$.003  & .586$\pm$.010  & .586$\pm$.020 & \uline{.596$\pm$.035}  & .555$\pm$.005 & .580$\pm$.006 & \textbf{.630$\pm$.014} &   8.79\% \\
  \multicolumn{2}{c|}{}                                        & .182$\pm$.011 & .187$\pm$.007 & .201$\pm$.003  & .276$\pm$.021  & .249$\pm$.052 & .228$\pm$.077  & .230$\pm$.019 & \uline{.282$\pm$.013} & \textbf{.290$\pm$.026} &  29.46\% \\
  \multicolumn{2}{c|}{}                                        & .691$\pm$.010 & .586$\pm$.006 & .641$\pm$.002  & .781$\pm$.019  & .554$\pm$.012 & .566$\pm$.025  & .756$\pm$.029 & \textbf{.803$\pm$.017} & \uline{.801$\pm$.009} &  20.66\% \\ \hline
  \multicolumn{2}{c|}{\multirow{3}{*}{\textit{balance-scale}}} & .599$\pm$.021 & .587$\pm$.008 & .578$\pm$.010  & .610$\pm$.006  & .586$\pm$.004 & .586$\pm$.004  & .611$\pm$.008 & \uline{.624$\pm$.004} & \textbf{.673$\pm$.009} &  12.77\% \\
  \multicolumn{2}{c|}{}                                        & .587$\pm$.034 & .649$\pm$.017 & .611$\pm$.014  & .572$\pm$.008  & \textbf{.665$\pm$.013} & \uline{.653$\pm$.007}  & .569$\pm$.009 & .584$\pm$.006 & \uline{.653$\pm$.015} &   7.13\% \\
  \multicolumn{2}{c|}{}                                        & .814$\pm$.016 & .722$\pm$.007 & .698$\pm$.014  & .821$\pm$.010  & .781$\pm$.011 & .790$\pm$.002  & .827$\pm$.006 & \uline{.840$\pm$.003} & \textbf{.891$\pm$.015} &  13.67\% \\ \hline
  \multicolumn{2}{c|}{\multirow{3}{*}{\textit{cmc}}}           & .431$\pm$.005 & .452$\pm$.006 & .456$\pm$.008  & \textbf{.489$\pm$.006}  & .476$\pm$.004 & \uline{.486$\pm$.009}  & .472$\pm$.010 & .484$\pm$.009 & .482$\pm$.003 &   3.05\% \\
  \multicolumn{2}{c|}{}                                        & .152$\pm$.006 & .192$\pm$.009 & .196$\pm$.014  & \uline{.243$\pm$.008}  & .224$\pm$.005 & .240$\pm$.012  & .218$\pm$.016 & .238$\pm$.014 & \textbf{.268$\pm$.005} &  28.69\% \\
  \multicolumn{2}{c|}{}                                        & .611$\pm$.003 & .627$\pm$.005 & .630$\pm$.002  & \uline{.678$\pm$.004}  & .669$\pm$.001 & .668$\pm$.006  & .654$\pm$.005 & .663$\pm$.006 & \textbf{.686$\pm$.010} &   5.64\% \\
  \bottomrule
  \end{tabular}
\end{table*}
  
{\bf Datasets.} 
To verify the effectiveness of \method in practical applications, we extend the experiments to real-world IL tasks from the UCI repository~\cite{Dua2019uci}.
These data are collected from different application domains including bioinformatics, sociodemographics, clinical medicine, etc.
To ensure a thorough assessment, these datasets vary widely in their properties.
Please refer to \tablename~\ref{tab:datasets} for detailed data statistics.

{\bf Evaluation protocol.}
For IL problems, classification accuracy is not a fair performance metric as it cannot reflect how the classifier works on minority classes.
Hence, unbiased evaluation criterias based on the number of true/false positive/negative prediction are usually used in IL.
For a comprehensive evaluation, we consider 3 unbiased metrics: macro-averaged F1-score, MCC (matthews correlation coefficient), and AUROC (area under the receiver operating characteristic curve)~\cite{hand2001auroc}.
For each dataset, we report the result of 5-fold stratified cross-validation and execute 5 independent runs with different random seeds to eliminate the randomness.

\subsubsection{Comparison with Resampling IL Methods}
\label{sec:exp-real-resampling}
We first compare \method with resampling-based IL solutions.
They have been widely used in practice for the preprocessing of class-imbalanced data~\cite{haixiang2017learning-from-imb-review}.
Ten representative methods are selected from 4 major branches of resampling-based IL: under-\&over-sampling and over-sampling with cleaning post-process (also referred as hybrid-sampling in some literature, we do not use this name to prevent confusion).
All methods are tested on the \textit{ozone-level} dataset, which has the highest imbalance ratio (IR=33.75), to test their efficiency and effectiveness.
The ensemble size of \method is set to 10, with SHEM and PBDA.
Five different classifiers, i.e., Multi-layer Perceptron (MLP), K-nearest neighbor (KNN), decision tree (DT), adaptive boosting (BST), and Bagging (BAG), are used to collaborate with these approaches.
The number of training samples and the time used to perform resampling are also reported to demonstrate the computational efficiency.

\tablename~\ref{tab:results-resamp} details the experiment results.
We show that by explicitly performing inter-\&intra-class balancing, \method outperforms canonical resampling methods by a significant margin.
In such a highly imbalanced dataset, minority class is likely to be poorly represented and lacks a clear structure.
Thus the advanced resampling methods that rely on distance-computing and neighborhood relations between minority objects may deteriorate the classification performance, especially when working with high-capacity models (e.g., BST and BAG).
The over-sampling + cleaning methods generally perform better than other baselines as they combine under-sampling (US) and over-sampling (OS). 
But such combination also introduces more distance computational overhead and makes the resampling time considerably high.
We also notice that distance-based US is usually more costly than OS as it involves calculating the distance between the majority and minority instances (e.g., {\sc Condense} resampling takes 12157.64 ms), while OS only considers the distance within the minority class.
In contrast, {\sc DuBE}'s resampling does not involve any distance calculation and is therefore computationally efficient.

\subsubsection{Comparison with Ensemble IL Methods}
\label{sec:exp-real-ensemble}

We further compare \method with 8 representative ensemble IL solutions on 10 real-world imbalanced classification tasks.
Baselines include 4 under-sampling-based methods ({\sc RusBst}~\cite{seiffert2010rusboost}, {\sc UnderBag}~\cite{barandela2003underbagging}, {\sc Cascade}~\cite{liu2009ee-bc}, and {\sc Spe}~\cite{liu2020self-paced-ensemble}), and 4 over-sampling-based ones ({\sc OverBst}~\cite{galar2012ensemble}, {\sc SmoteBst}~\cite{chawla2003smoteboost}, {\sc OverBag}~\cite{galar2012ensemble}, and {\sc SmoteBag}~\cite{wang2009smotebagging}).
"{\sc Bst}"/"{\sc Bag}" indicates that the method is based on Adaptive Boosting/Bootstrap Aggregating (boosting/bagging) ensemble learning framework.
For a fair comparison, the ensemble size of all test methods is set to 10.
\method is implemented with RHS, SHEM and PBDA.
We use C4.5 decision tree as the base learner for all ensembles following the setting of most of the previous works~\cite{haixiang2017learning-from-imb-review}.

The results are reported in \tablename~\ref{tab:results-ens}.
We also report {\sc DuBE}'s average performance improvement relative to all baselines ($\Delta_\text{mean}$) in percentage.
It can be observed that \method achieves competitive performance in various real-world IL tasks.
It outperforms all other 8 ensemble methods in 26 out of 10$\times$3 task-metric pairs.
On average, \method brings significant performance improvements over existing ensemble IL baselines (1.98\%/15.22\%/39.82\% min/ave/max $\Delta_\text{mean}$) by explicitly considering inter- and intra-class imbalance.
We can also see that bagging-based approaches generally perform better than boosting-based ones (e.g., on all 30 task-metric pairs, the performance of {\sc SmoteBag} is 6.54\% better than {\sc SmoteBst}).
Compared with boosing that only manipulates sample weights, the bootstrap sampling in bagging introduces additional data-level diversity, which helps prevent overfitting in IL, especially on minority class(es).
Note that boosting is a way to implement HEM by reweighting. 
On the other hand, {\sc Cascade} and {\sc Spe} also incorporate the idea of HEM, but by resampling.
We find that the latter two resampling-based methods also usually outperform the boosting-based ones (mean({\sc Cascade}, {\sc Spe}) is 10.87\% better than mean({\sc RusBst, OverBst, SmoteBst})), which further validates the importance of data-level diversity in IL.
In DuBE, in addition to reducing decision bias, RHS and PBDA also play important roles in diversifying the base classifiers and preventing overfitting.

\subsubsection{Ablation Study and Discussions}
\label{sec:ablation}
We carry out further experiment to validate the contribution of different inter-class (RUS/ROS/RHS) and intra-class balancing (HEM/SHEM) as well as the perturbation-based data augmentation (PBDA).

\begin{figure}[h]
  \centering
  \includegraphics[width=\linewidth]{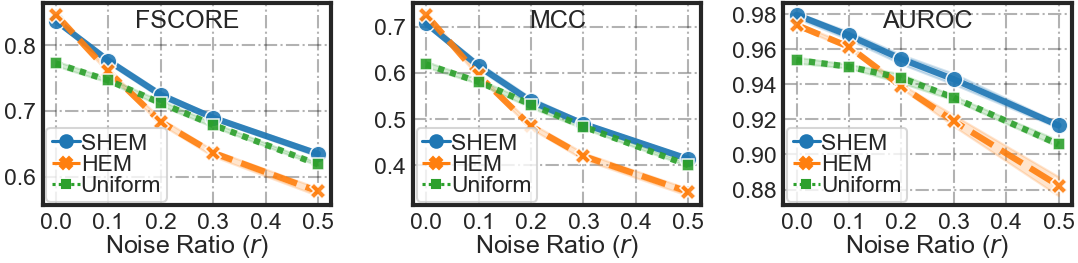}
  \caption{
  Comparison of IntraCB strategies under varying noise ratio.
  }
  \label{fig:intra-ablation}
\end{figure}

\textbf{Intra-class balancing (IntraCB).}
As previously discussed in \S~\ref{sec:intracb}, HEM helps alleviate intra-class imbalance by focusing on informative high-error instances, but is vulnerable to noise/outliers.
SHEM is thus designed for robust HEM.
In this experiment, we test their robustness against noise.
In practice, the training data often contains corrupted labels due to errors in the labeling process like crowdsourcing.
Here we simulate this by introducing flip noise, i.e., with noise ratio $r$, $r\cdot n_\text{minority}$ minority samples will be assigned opposite labels and vice versa.
The largest \textit{letter-img} dataset is used as a representative.
We test {\sc DuBE}$_\text{10,RUS}$ with HEM/SHEM and without IntraCB (uniform) on the 0\%-50\% corrupted data, as shown in \figurename~\ref{fig:intra-ablation}.
We can observe that when the data contains no noise ($r$=0), both HEM and SHEM achieve significantly better performance than uniform sampling.
However, HEM's performance drops rapidly as $r$ increases, and is even worse than uniform resampling when $r\geq 0.2$.
In contrast, SHEM is more robust and consistently outperforms uniform resampling at different noise levels, which validates its effectiveness.

\begin{figure}[h]
  \centering
  \includegraphics[width=\linewidth]{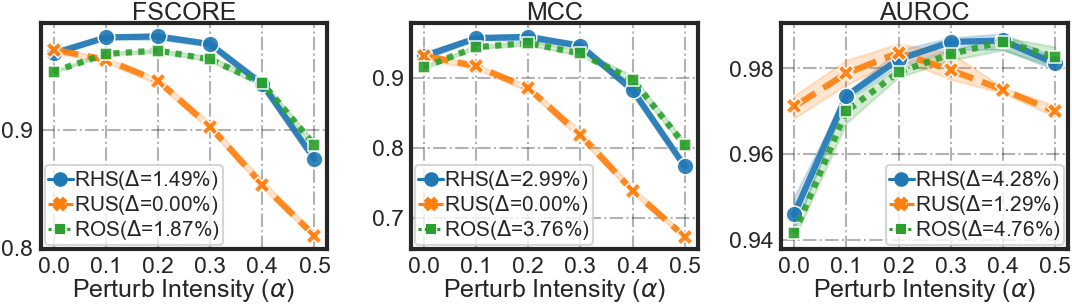}
  \caption{
  Comparison of InterCB strategies under varying perturb intensity $\alpha$, $\Delta$ is the relative performance gain from PBDA, i.e.,
  $\frac{\text{max}(\text{score}_{\alpha>0}) - \text{score}_{\alpha=0}}{\text{score}_{\alpha=0}}$.
  }
  \label{fig:inter-ablation}
\end{figure}

\textbf{Inter-class balancing (InterCB) and PBDA.}
In \S~\ref{sec:intercb}, we have discussed the difference between RUS, RHS and ROS.
Specifically, we find that RUS is very effective in mitigating decision bias as it can be seen as an unbiased correction to the skewed marginal distribution $P(Y)$.
The replication-based over-sampling in RHS and ROS on the other hand, is not equivalent to resampling from the underlying distribution $P(X|Y=c_\text{min})$ and thus does not help to mitigate bias.
But recall that in \S~\ref{sec:pbda}, we further introduce PBDA, which could be used to improve RHS and ROS by augmenting duplicated minority instances.
To confirm the effectiveness of different InterCB strategies and PBDA, we test {\sc DuBE}$_\text{10,SHEM}$ with RUS, RHS, ROS and different perturb intensity $\alpha$.
Results are shown in \figurename~\ref{fig:inter-ablation}.
It can be observed that in the absence of PBDA ($\alpha$ = 0), pure RUS outperforms RHS and ROS, especially in terms of AUROC scores, because of its ability to naturally mitigate bias.
However, we also notice that RUS has significantly weaker performance gains from PBDA compared with RHS and ROS, as indicated by $\Delta$s in \figurename~\ref{fig:inter-ablation}.
With proper data augmentation, RHS\&ROS achieve better performance.
These findings are consistent with our discussions in \S~\ref{sec:pbda} \& \S~\ref{sec:intercb}.
We also find that metrics have different responses to PBDA, e.g., for RHS, the optimal $\alpha$ = 0.2 for F1-score and MCC, but 0.4 for AUROC.

\begin{figure}[h]
  \centering
  \includegraphics[width=\linewidth]{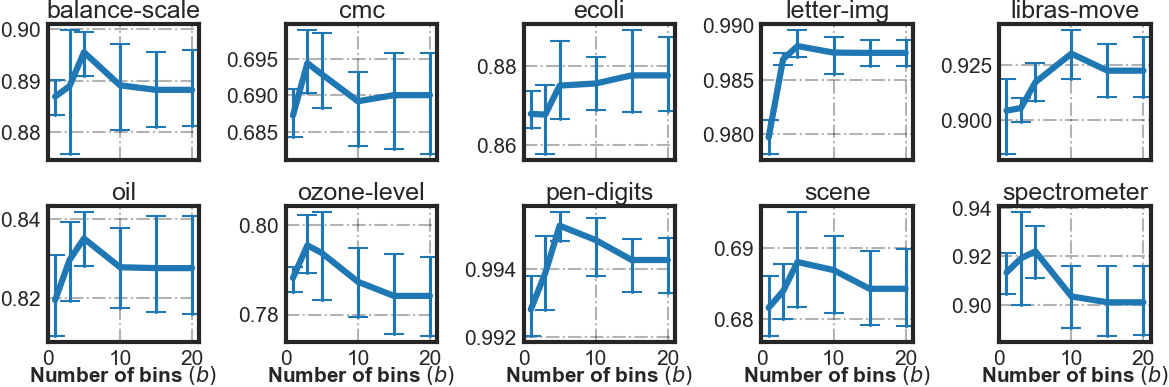}
  \caption{
    The influence of the number of bins $b$ (macro AUROC).
  }
  \label{fig:bins}
\end{figure}

\begin{figure}[h]
  \centering
  \includegraphics[width=\linewidth]{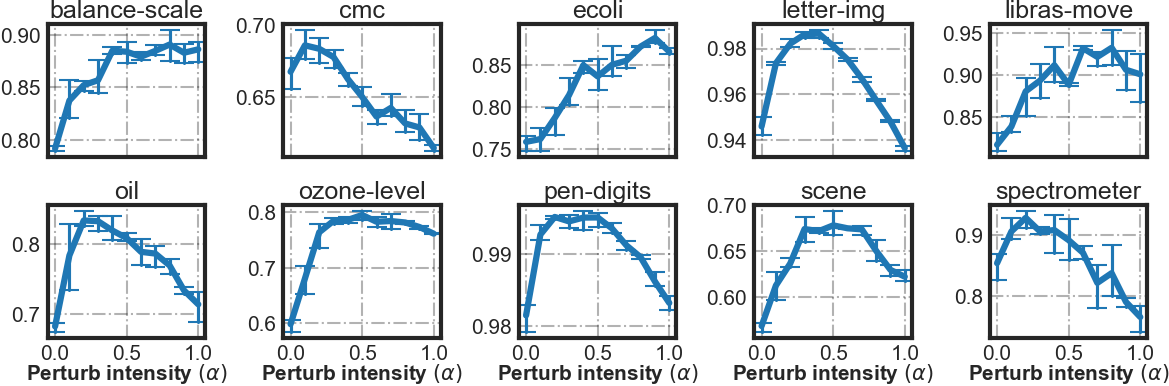}
  \caption{
    The influence of the perturbation coefficient $\alpha$ (macro AUROC).
  }
  \label{fig:alpha}
\end{figure}

\textbf{Parameter analysis.}
\method has two main parameters: the number of bins in the histogram $b$ and the perturbation coefficient $\alpha$.
In this section, we discuss their influence based on real-world IL tasks.
First, $b$ determines how detailed the error distribution approximation is, we show its influence in \figurename~\ref{fig:bins}.
We can see that setting a small $b$ may lead to a poor performance, e.g., $b$=1 degrades SHEM to uniform sampling.
Using a large $b$ (e.g., $\ge10$) does not necessarily lead to better performance either, as many bins may be empty.
For these reasons, we recommend setting the $b$ to be 5, which is also the setting we used in the experiments. 
One can try increasing $b$ when working on large datasets.
On the other hand, $\alpha$ determines the intensity of the perturbation-based augmentation.
\figurename~\ref{fig:alpha} shows it influence in real-world tasks.
We can see that with proper $\alpha$, PBDA significantly helps generalization, but keep increasing $\alpha$ will introduce too much perturbation and degrade the performance.
In our implementation of {\sc DuBE}, $\alpha$ could be automatically tuned using a small validation subset.

\textbf{Implementation details.}
Our implementation of \method is based on Python 3.8.5.
We use open-source software packages \textit{imbalanced-learn}~\cite{guillaume2017imblearn} and \textit{imbalanced-ensemble}~\cite{liu2021imbens} for implementation of all baseline resampling and ensemble IL methods.
The base learning models (MLP, KNN, DT, etc.) are from the \textit{scikit-learn}~\cite{pedregosa2011sklearn} package.
The resampling times are obtained based on the results of running on an Intel Core$^\text{TM}$ i7-10700KF CPU with 32GB RAM.

\textbf{Complexity analysis.}
The complexity of \method (Alg.\ref{alg:dube}) mainly comes from the intra-class balancing (line\#8), which requires the latest prediction probabilities estimated by the current ensemble.
Suppose the complexity for a base classifier $f(\cdot)$ to predict on dataset $D$ is $C^{pred}_{f, D}$.
Then the total resampling cost for training a $k$-classifier \method is $k^\text{pred}\cdot C^{pred}_{f, D}$, where $k^\text{pred}$ is the number of times of making predictions with $f(\cdot)$.
Normally we have $k^\text{pred}=\Sigma_{t=1}^{k-1} t = \frac{k(k-1)}{2}$ according to Alg.\ref{alg:dube}, i.e., resampling complexity is $O(k^2 C^{pred}_{f, D})$.
But as \method is an additive ensemble, we can reduce $k^\text{pred}$ to $k$ by buffering the predicted probabilities (a $N\times m$ matrix) for each fitted $f(\cdot)$.
Together with the training cost, the final complexity of DuBE is $O(k(C^{pred}_{f,D}+C^{train}_{f, D'}))$, note that the $C^{train}_{f, D'}$ term depends on the InterCB strategy used, since $|D'_\text{RUS}|$<$|D'_\text{RHS}|$<$|D'_\text{ROS}|$.
Finally, we review the related works.
\section{Related Works}
\label{sec:related-works}

He et al.~\cite{he2008overview,he2013overview}, Guo et al.~\cite{haixiang2017learning-from-imb-review} and Krawczyk et al.~\cite{krawczyk2016learning} provided systematic surveys of algorithms and applications of class-imbalanced learning.
In this section, we review the research topics that are closely related to this paper.

\subsection{Class-wise Re-balancing.}
As mentioned earlier, the vast majority of existing IL solutions fall into this category.
They can be divided into two regimes: resampling and reweighting.
{\it Resampling} methods focus on directly modifying the training set to balance the class distribution (e.g., over-sampling~\cite{chawla2002smote,he2008adasyn,han2005borderline-smote} and under-sampling~\cite{mani2003nearmiss,tomek1976tomeklink,wilson1972enn,kubat1997oss}).
Beyond the na\"ive solutions, previous efforts have adopted different heuristics to guide their resampling process.
For example, \cite{han2005borderline-smote,mani2003nearmiss} aim to generate or keep instances that are close to the borderline/overlap area.
But on the contrary, \cite{wilson1972enn,batista2004smoteenn,ramentol2012smotersb} consider examples located in the overlap zone to be detrimental to learning, and therefore discard them from the trainig set.
Such methods often rely on exploring distance-based neighborhood information, e.g., {\sc Smote} over-sampling~\cite{chawla2002smote} and its variants~\cite{he2008adasyn,han2005borderline-smote,kovacs2019smotevari}, many denoising under-sampling techniques~\cite{tomek1976tomeklink,wilson1972enn,kubat1997oss}, as well as their hybrid usages~\cite{batista2004smoteenn,ramentol2012smotersb,batista2003smotetomek}.
On the other hand, \textit{Reweighting} approaches assign different misclassification costs to different classes (e.g., cost-sensitive learning~\cite{ling2004csdt,chai2004csnb}), thus forcing the model to focus on the pattern of minority classes.

However, these methods do not explicitly account for inter-class and intra-class imbalances in their design, and suffer from unsatisfactory performance, high computational cost, and poor applicability.
Distance-based resampling requires a well-defined distance metric, which may not be available in practice since the data may contain categorical features and missing values~\cite{liu2020self-paced-ensemble}.
Moreover, some of these algorithms run extremely slow on large-scale datasets as the cost of calculating the distance between each instances grows quadratically with the size of the dataset.
Reweighting, i.e., cost-sensitive learning, often requires targeted modifications to the learning algorithms~\cite{haixiang2017learning-from-imb-review}. 
More importantly, in most cases, it is difficult to obtain an appropriate cost matrix given by domain experts~\cite{krawczyk2014cost-ensemble}.

\subsection{Hard Example Mining.}
Most of the works in this group are raised in recent years with the boom in deep learning approaches, especially in computer vision applications such as image classification and object detection~\cite{johnson2019review-deep,oksuz2020review-object}.
They (implicitly) reweight the gradient update of samples based on their difficulties or losses, i.e., down-weight the well-classified examples and assign more weights to hard examples.
Such hard example mining (HEM) methods have achieved notable success in many areas.
For example, \textit{Qi et al.}~\cite{dong2017rectification-hem} propose Class Rectification Hard Mining to handle imbalanced image classification.
To deal with the background-foreground imbalance in object detection, Online HEM~\cite{shrivastava2016ohem} only back-propagates gradients for hard examples in the later phase of training. 
\textit{Lin et al.}~\cite{lin2017focalloss} further propose a uniform HEM loss function FocalLoss for object detection.

We note that there are some inter-class balancing solutions that that also incorporate the idea of HEM.
For instance, \textit{Liu et al.}~\cite{liu2009ee-bc} iteratively discard well-classified majority samples when perform under-sampling during ensemble training.
The IL methods (e.g.,~\cite{chawla2003smoteboost,seiffert2010rusboost,sun2007cost-boost}) that based on adaptive boosting (AdaBoost) can also be considered as incorporating HEM, as AdaBoost emphasizes those instances misclassified by previous classifiers.
Nevertheless, this does not mean that HEM is the optimal solution for intra-class balancing.
It may work well when the data is clean with no presence of noise.
But on a noisy dataset, most of hard examples are likely to be outliers.
They will be wrongly reinforced by direct HEM, which degrades the generalization performance.
Therefore, an adaptive approach is needed to achieve more robust intra-class balancing.

\subsection{Data Complexity Factors.}
Several research efforts have discussed the data complexity factors present in class-imbalanced learning.
These factors are believed to be widely existed in imbalanced datasets and contribute to the poor classification performance.
Specifically, \textit{Garcia et al.}~\cite{garcia2015noise} and \textit{Koziarski et al.}~\cite{koziarski2020multi-ccr} discussed the influence of noise examples in imbalanced learning.
Besides, \cite{napierala2016types-minority-samples,denil2010overlap,garcia2007overlap} suggest that class overlap/separability is the major problem responsible for degradation of classifier’s performance.
Some other works focus on the problem of small disjuncts~\cite{prati2004overlap-small-disjuncts,jo2004small-disjuncts}, i.e., presence of small-sized clusters (called subconcepts) containing examples from one class and located in the region of another class.
We note, however, that these data complexity factors are closely related to each other: severe class overlap can yields more outliers who play a similar role to noise examples~\cite{gupta2018overlap-noise}, and an increase in the amount of noise could also induce more small disjuncts~\cite{garcia2015noise-small-disjuncts}, and vice versa.
Compare with them, the intra-class imbalance introduced in this paper is a higher-level indicator of \textit{task complexity}.
It considers the data distribution in terms of learning difficulty (with respect to a given model), which reflects the ultimate effect of these data complexities on the classifier learning process.

\subsection{Ensemble Imbalanced Learning.}
By merging the outputs of multiple classifiers, ensemble imbalanced learning (EIL) is known to effectively improve typical IL solutions (e.g.,~\cite{chawla2003smoteboost,liu2009ee-bc,liu2020self-paced-ensemble,seiffert2010rusboost,wang2009smotebagging}).
These EIL solutions are gaining increasing popularity~\cite{haixiang2017learning-from-imb-review} and have demonstrated competitive performance in many IL tasks~\cite{krawczyk2016learning}.
But most of them are direct combinations of a resampling/reweighting scheme and an ensemble learning framework, e.g., {\sc Smote~\cite{chawla2002smote}+AdaBoost~\cite{freund1997adaboost}=SmoteBoost~\cite{chawla2003smoteboost}}.
It means that these methods inherit the same shortcomings of existing inter-class balancing strategies, such as the reliance on neighborhood relationships based on distance computing.
Consequently, albeit EIL techniques effectively lower the variance introduced by resampling/reweighting, they do not essentially improve the balancing strategy and thus still leave much room for improvement.
Note that DuBE is also a generic iterative ensemble learning framework, it takes the advantage of both classifier ensemble and the iterative training process that enables hard example mining.
\section{Conclusion \& Limitations}
\label{sec:conclusion}

In this paper, we discuss two types of imbalance that existed in the nature of the IL, i.e., inter- and intra-class imbalance, and how they implicitly correspond to existing IL strategies.
To explicitly handle them in a unified learning framework, \method is proposed, along with a systematic discussion on the pros and cons of exisitng Inter/IntraCB solutions.
Extensive comparative studies validate the effectiveness of {\sc DuBE}.
Beyond the discussions in this paper, we believe that more in-depth theoretical analysis is needed to find better solutions for inter-\&intra-class balancing.
To summarize, this work contributes preliminary effort towards understanding the inter-class and intra-class imbalance in IL, 
and we hope this work can shed some light on finding new IL solutions.


\section*{Acknowledgments}
This work is supported by the National Natural Science Foundation of China under Grant No.61976102, No.U19A2065,
and Science and Technology Development Program of Jilin Province under Grant No. 20210508060RQ.

\ifCLASSOPTIONcaptionsoff
  \newpage
\fi



%



\bibliographystyle{IEEEtran}
\bibliography{main}

%

\begin{IEEEbiography}[{\includegraphics[width=1in,height=1.25in,clip,keepaspectratio]{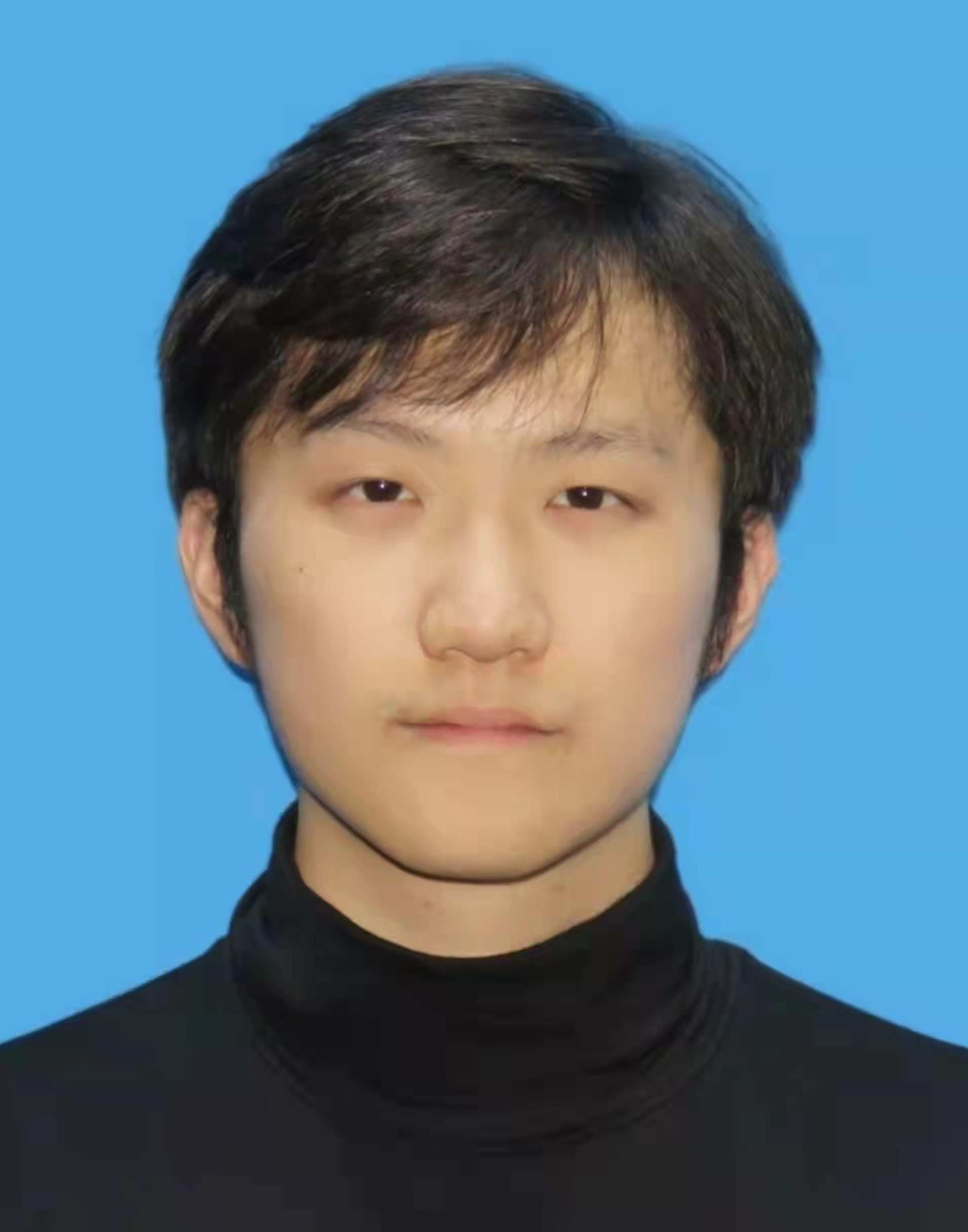}}]{Zhining Liu}
  is currently a Ph.D. student with the Department of Computer Science, University of Illinois Urbana-Champaign, USA.
  He received his B.S. and M.Eng. in Computer Science from Jilin University in 2019 and 2022, respectively.
  His recent research interests include imbalanced learning, ensemble learning, and meta-learning.
  He has published several research papers in the field of data mining in premium international conferences such as ICDE and NeurIPS.
\end{IEEEbiography}

\begin{IEEEbiography}[{\includegraphics[width=1in,height=1.25in,clip,keepaspectratio]{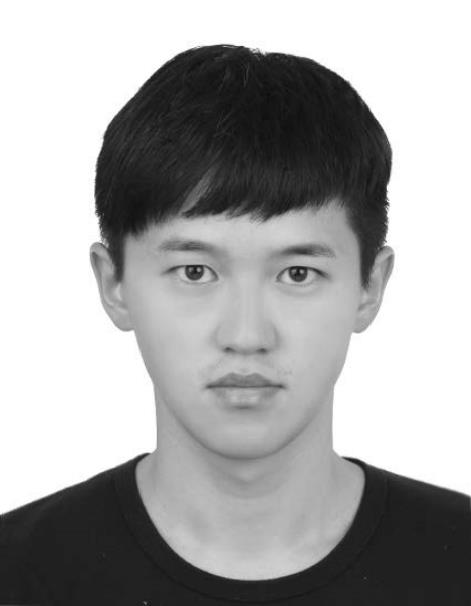}}]{Pengfei Wei}
  received the Ph.D. degree from the School of Computer Science and Engineering, Nanyang Technological University, Singapore, in 2019. 
  He is currently a Research Scientist with Bytedance AI Lab, Singapore. 
  His research interests include data analysis, transfer learning, domain adaptation, and reinforcement learning. 
\end{IEEEbiography}

\begin{IEEEbiography}[{\includegraphics[width=1in,height=1.25in,clip,keepaspectratio]{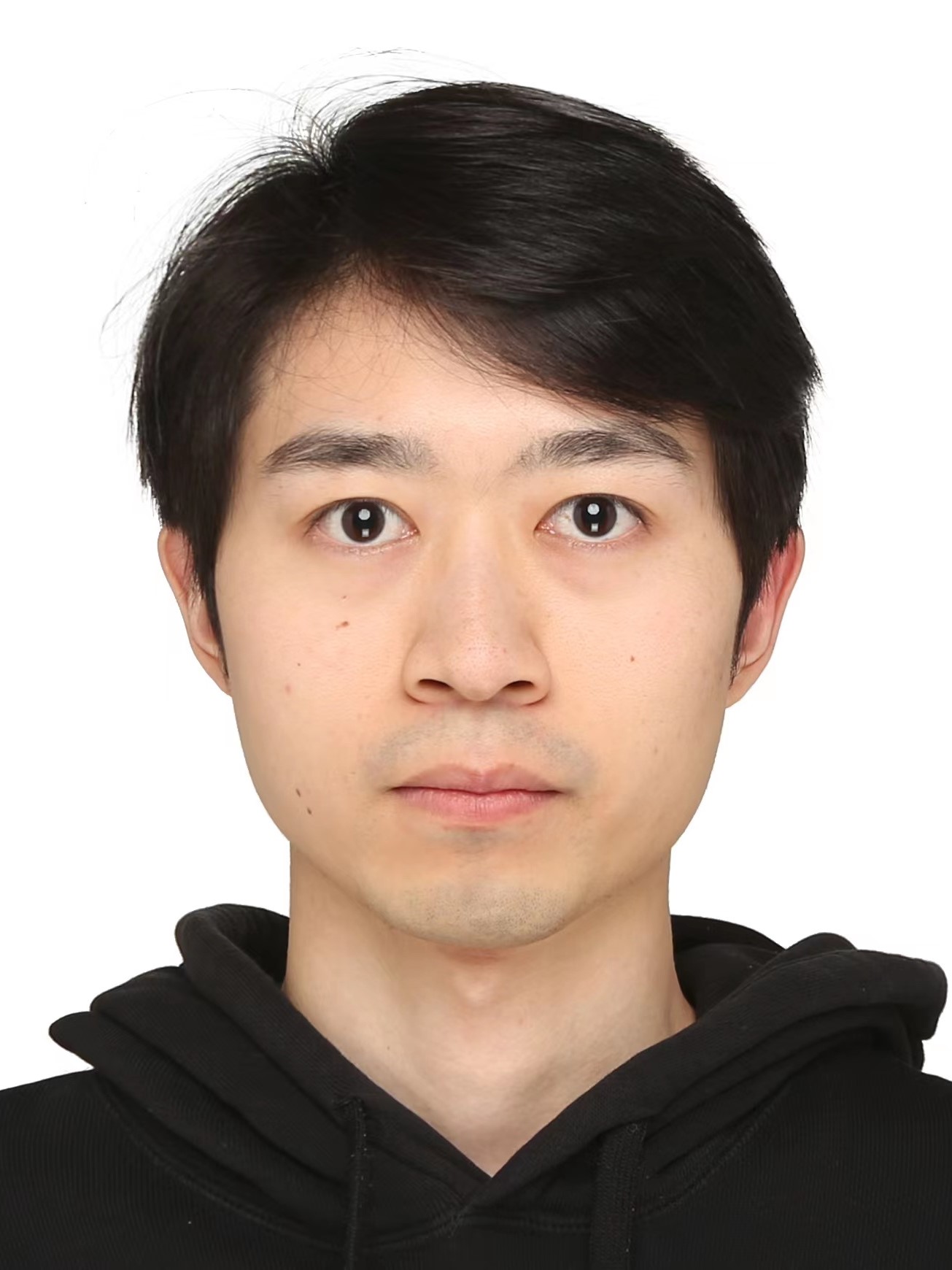}}]{Zhepei Wei}
  is currently a Ph.D. student with the Department of Computer Science, University of Virginia, USA.
  He received his B.S. and M.Eng. in Computer Science from Jilin University, China, in 2019 and 2022, respectively.
  His research interests include machine learning, data mining, and natural language processing.
\end{IEEEbiography}

\begin{IEEEbiography}[{\includegraphics[width=1in,height=1.25in,clip,keepaspectratio]{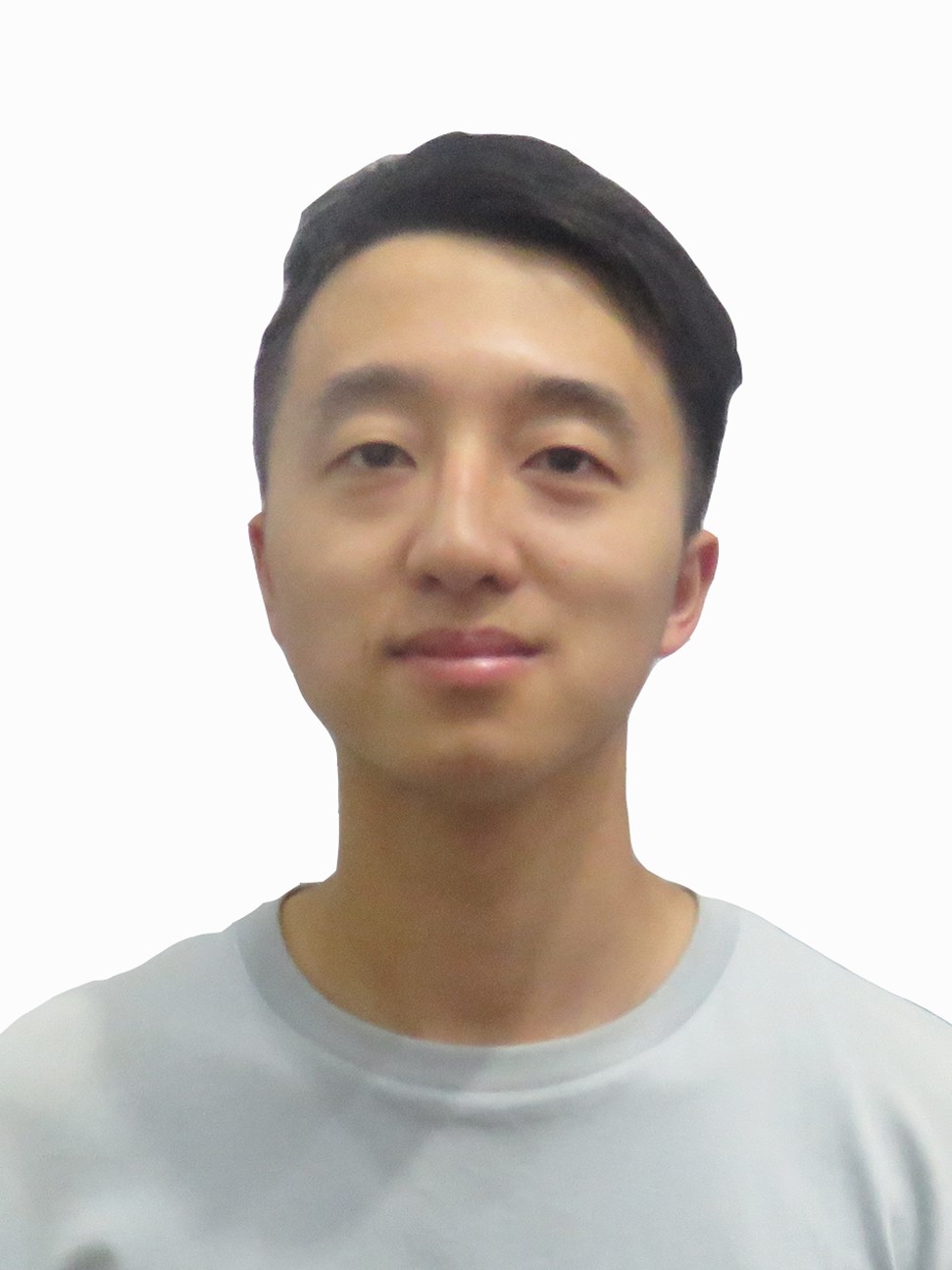}}]{Boyang Yu}
  is currently an Algorithm Engineer at JD.com.
  He received his B.S. and M.Eng. in Computer Science from Jilin University, China, in 2019 and 2022, respectively.
  His research interests include machine learning, data mining, and reinforcement learning.
\end{IEEEbiography}

\begin{IEEEbiography}[{\includegraphics[width=1in,height=1.25in,clip,keepaspectratio]{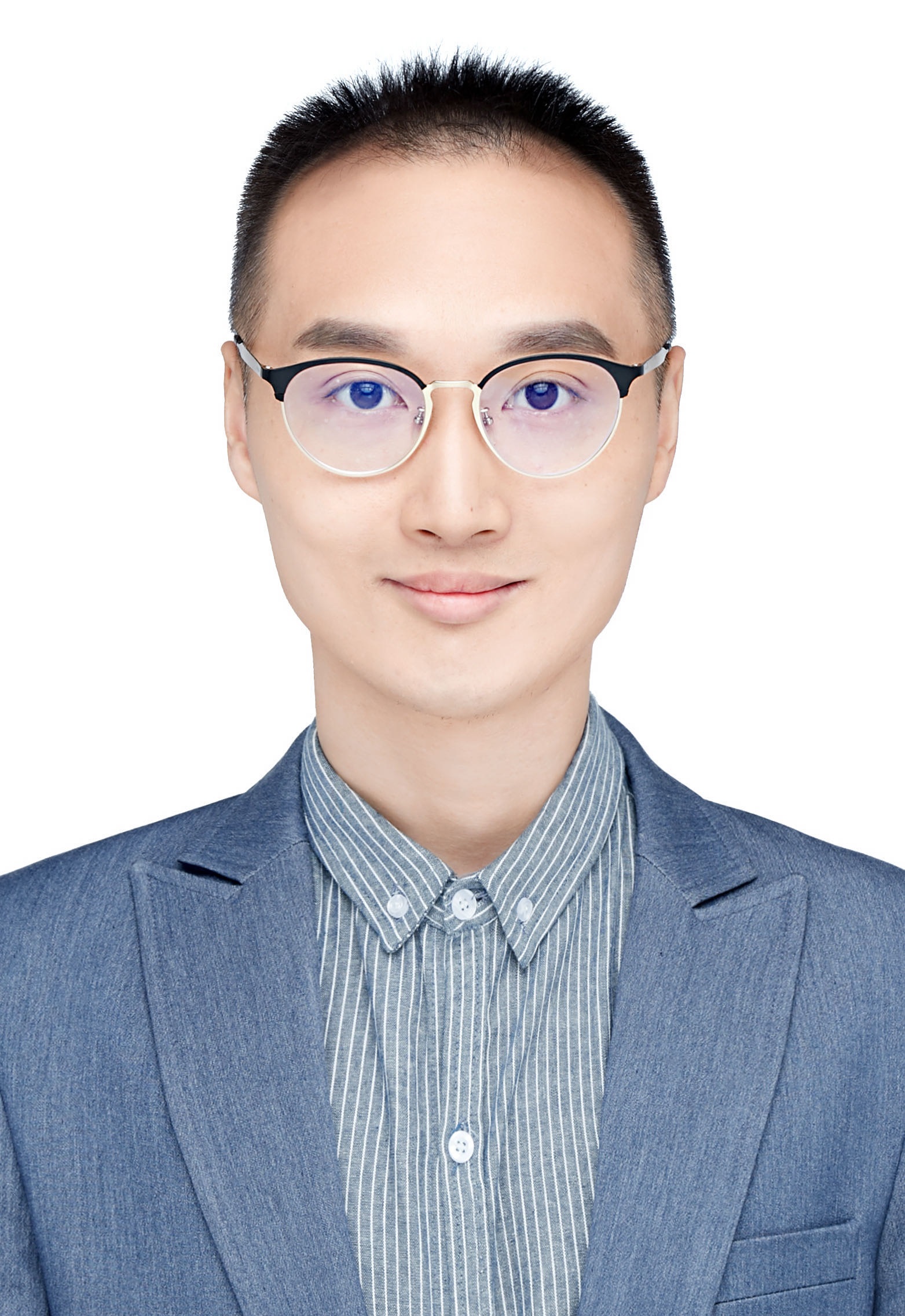}}]{Yuan Tian}
  received the Ph.D. degree from Jilin University in 2018.
  He is currently an Associate Professor with School of Artificial Intelligence, Jilin University, China.
  His research interests include machine learning, data mining, and natural language processing.
\end{IEEEbiography}

\begin{IEEEbiography}[{\includegraphics[width=1in,height=1.25in,clip,keepaspectratio]{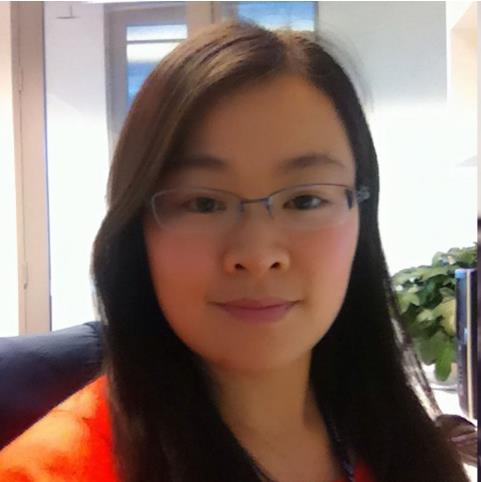}}]{Jing Jiang}
  received her PhD degree from the University of Technology Sydney (UTS), Australia in 2015. 
  She is currently a Lecturer at the Centre for Artificial Intelligence, Faculty of Engineering and IT, UTS. 
  Her research interest lies in data mining and machine learning applications with the focuses on deep reinforcement learning and sequential decision-making. 
  She has more than 20 research papers published on top-tier journals and conferences including AAAI, IJCAI and CCGrid. 
\end{IEEEbiography}

\begin{IEEEbiographynophoto}{Wei Cao}
  received the Ph.D. degree from Institute for Interdisciplinary Information Sciences (IIIS) at Tsinghua University in 2018.
  He is currently a senior researcher at Microsoft Research Asia (MSRA), Machine Learning Group.
  He has a broad research interests on applied machine learning, including Time Series Forecasting, Intelligent Health-Care, Logistics Optimization etc.
\end{IEEEbiographynophoto}

\begin{IEEEbiography}[{\includegraphics[width=1in,height=1.25in,clip,keepaspectratio]{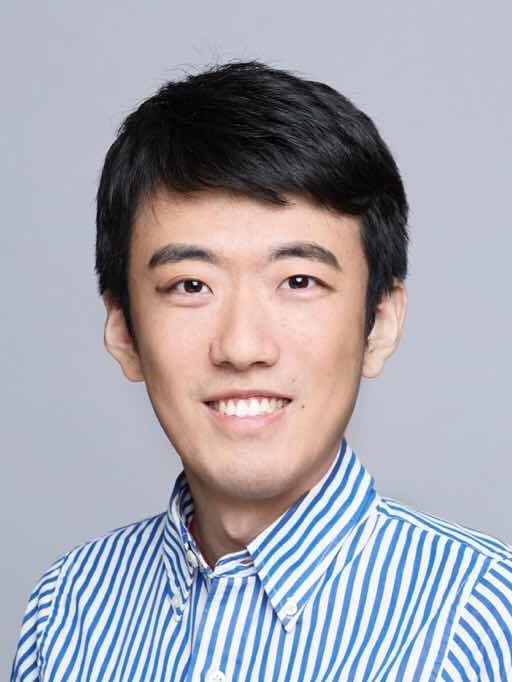}}]{Jiang Bian}
  received the Ph.D. degree from Georgia Institute of Technology in 2010.
  He is currently a Researcher and Engineer with rich experience in information retrieval, data mining, and machine learning. 
  He is also a Principal Researcher and a Research Manager with Microsoft Research, Beijing, China, with research interests in AI for finance, AI for logistics, deep learning, multiagent reinforcement learning, computational advertising, and a variety of machine learning applications. 
\end{IEEEbiography}

\begin{IEEEbiography}[{\includegraphics[width=1in,height=1.25in,clip,keepaspectratio]{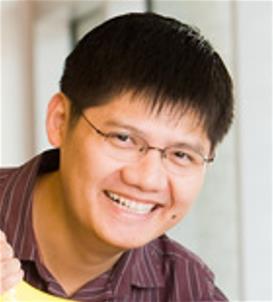}}]{Yi Chang}
is the Dean of the School of Artificial Intelligence, Jilin University. 
His research interests include information retrieval, data mining, machine learning, natural language processing, and artificial intelligence.
He has published more than 100 research papers in premium conferences or journals.
He served as an associate editor of IEEE TKDE, and one of the conference General Chairs for ACM WSDM'2018 and ACM SIGIR'2020. 
He is an IEEE Senior Member and ACM Distinguished Scientist.
\end{IEEEbiography}







\end{document}